%% file: main.tex
\documentclass[11pt]{article}

\usepackage[final]{acl}

\usepackage{times}
\usepackage{latexsym}
\usepackage[T1]{fontenc}
\usepackage[utf8]{inputenc}
\usepackage{microtype}
\usepackage{hyperref}
\usepackage{url}
\usepackage{booktabs}
\usepackage{xspace}
\usepackage{graphicx}
\usepackage{textcomp}
\usepackage{multicol}
\usepackage{multirow}
\usepackage{enumitem}
\usepackage{float}
\usepackage{tikz}
\usepackage{subcaption}
\usepackage{makecell}
\usepackage{CJKutf8}
\usepackage{inconsolata}
\usepackage{adjustbox}
\usepackage{amssymb}
\usepackage[dvipsnames,table]{xcolor}
\usepackage{pifont}
\usepackage{algorithm}
\usepackage[noEnd=false,indLines=true]{algpseudocodex}
\usepackage{amsmath}
\definecolor{Gray2}{gray}{0.45}
\usepackage[most]{tcolorbox}
\usepackage{soul}  
\usepackage{tabularx}

\definecolor{c1}{RGB}{200,225,250}  
\definecolor{c2}{RGB}{210,240,210}  
\definecolor{c3}{RGB}{255,230,200}   

\newcommand{\hlA}[1]{{\sethlcolor{c1}\hl{#1}}}
\newcommand{\hlB}[1]{{\sethlcolor{c2}\hl{#1}}}
\newcommand{\hlC}[1]{{\sethlcolor{c3}\hl{#1}}}
\newcommand{\violate}[1]{\textcolor{red}{#1}}

\usepackage{placeins}

\definecolor{darkblue}{rgb}{0, 0, 0.5}
\hypersetup{colorlinks=true, citecolor=darkblue, linkcolor=darkblue, urlcolor=darkblue}

\definecolor{mycolor}{RGB}{33, 95, 154}
\definecolor{custom_red}{RGB}{228, 54, 54}

\title{SyRuP: Enhancing System-Prompt Following \\ via Reward-Guided Prediction in LLM Decoding}

\def \name{\textsc{SyRuP}\xspace}

\author{
  Seoyeon Kim \qquad
  Minjae Kang \qquad
  Jaehyung Kim \\
  Yonsei University \\
  \texttt{\{yeon\_04, mjkang618, jaehyungk\}@yonsei.ac.kr}
}

\begin{document}
\maketitle

\input{0_abstract}
\input{1_intro}
\input{2_related}
\input{3_method}
\input{4_experiment}
\input{5_conclusion}

\bibliography{custom}
\clearpage
\appendix
\input{6_appendix}

\end{document}

%% file: 0_abstract.tex
\begin{abstract}

Large Language Models (LLMs) are increasingly controlled through system prompts that specify roles, formats, and safety requirements.
However, models follow these prompts only implicitly through in-context learning, which can be insufficient for complex or compositional prompts.
Existing approaches often require model tuning or response-level reranking, limiting their practicality for lightweight inference-time control.
We introduce \name{}, a decoding-time framework for improving system-prompt adherence while keeping the base LM frozen.
\name{} trains a cross-attention reward head from system-prompt-conditioned preference pairs, treating the system prompt as a separate memory to produce token-level adherence scores.
At inference, \name{} reranks the base LM's top-$k$ candidates by combining base logits with both the learned reward signal and a contrastive signal that captures system-induced logit shifts.
Experiments on system-prompt following benchmarks show that \name{} consistently outperforms prompting and decoding-time baselines with moderate inference overhead.
These results suggest that explicit token-level guidance is an effective and practical mechanism for reliable system-prompt following.\looseness=-1\footnote{The code is available at \url{https://github.com/ssseoyneee/EMNLP26_SyRuP}.}
\end{abstract}

%% file: 1_intro.tex
\section{Introduction}
\input{Figures/Figure1}
\input{Figures/Figure2}

Large language models (LLMs) are increasingly deployed across diverse domains, where system prompts have become a standard mechanism for specifying personas, styles, constraints, and safety guidelines \citep{zhang2025iheval, lee2024aligning, wang2024rnr, wallace2024instruction}. 
Unlike user inputs, system prompts are typically set in advance and serve as persistent, high-priority instructions that govern model behavior throughout an interaction.
However, at inference time, LLMs must follow a given system prompt solely through conditioning, without explicit optimization for that particular prompt. 
As system prompts become increasingly complex and compositional, the models often fail to adhere to them consistently \citep{qin2024sysbench, mu2025closer}.
While training-time approaches such as SFT and RLHF can improve general instruction-following behavior, they require costly parameter updates and may incur alignment tax \citep{ouyang2022training,lin2024mitigating}.
Decoding-time approaches avoid these limitations by steering frozen models during generation, but have primarily focused on generic instruction following rather than the explicit evaluation and steering of system-prompt adherence \citep{kang2026enhancing, liu2024decoding, mudgal2023controlled}.

Among decoding-time approaches, \textit{reward-guided decoding} offers a promising direction by using a learned reward model to modulate next-token probabilities during generation \citep{khanov2024args}. 
However, reward models typically process the prompt and response as a single sequence, causing the system prompt to be absorbed into the prefix context rather than treated as a separate conditioning source.
Moreover, prior methods have mainly focused on fixed and task-level attributes, such as preference alignment, detoxification, and sentiment control \citep{deng2023reward, kim2023critic}, which limits their applicability to system-prompt following, where constraints are more diverse and adherence must be judged relative to each instance. 
A parallel line of work, \textit{contrastive decoding}, exploits the distributional difference induced by the presence or absence of the context \citep{li2023contrastive, shi2024trusting, leng2024mitigating}.
This training-free signal is naturally applicable to the system prompt, where it captures the base LM's intrinsic sensitivity to the given instruction. 
However, it remains implicit: it does not provide an explicit, learned judgment of whether a candidate token satisfies the prompt.
Consequently, a reward-guided decoding framework that treats the system prompt as a separate conditioning source and explicitly evaluates token-level adherence to it remains underexplored, as does the question of how this explicit reward signal can be effectively strengthened by the LM's intrinsic contrastive signal.\looseness=-1

We therefore propose \textbf{\name{}} (\textbf{Sy}stem-prompt following via \textbf{R}eward-g\textbf{u}ided \textbf{P}rediction), a decoding-time guidance method that keeps the backbone LM frozen and steers generation with a learned reward signal, reinforced by an intrinsic contrastive signal.
We reformulate system-prompt following not as simple prefix conditioning, but as a token-level conditional adherence problem between a fixed system instruction and an evolving generation state.
To this end, we train a lightweight cross-attention reward head that uses the system-prompt hidden states as key-value memory.
Specifically, it constructs query states by concatenating the last-layer hidden states of the user query and generated prefix, and uses the final position of the cross-attended sequence to produce token-level rewards; at each generation step, the reward head produces an explicit adherence score for each candidate token.
To complement this learned judgment, \name{} further incorporates a contrastive signal that captures the system-prompt-induced shift in the base LM's token distribution.
The two signals are combined with the frozen base LM logits to rerank candidate tokens.

We demonstrate the effectiveness of \name{} on Multifaceted-Bench \citep{lee2024aligning} and IFEval \citep{zhou2023instruction}, using three backbone LLMs and a range of prompting-based and decoding-time baselines.
On Multifaceted-Bench, \name{} achieves the best average score for every backbone, improving over the strongest baseline by 5.58\% on \texttt{Qwen2.5-1.5B-Instruct}, and ranks first across all 15 model-task combinations.
\name{} also achieves performance comparable to the strongest baselines on IFEval without task-specific training, demonstrating its generalization beyond the benchmark family used for training.
Further analyses show that learned reward guidance and contrastive correction provide complementary token-level signals, and that their combination offers a favorable trade-off between system-prompt adherence and inference cost.
Overall, these results validate \name{} as an effective inference-time framework for improving system-prompt following while keeping the base LLM frozen.

%% file: Figures/Figure1.tex
\begin{figure}[ht]
\centering
\includegraphics[width=\columnwidth]{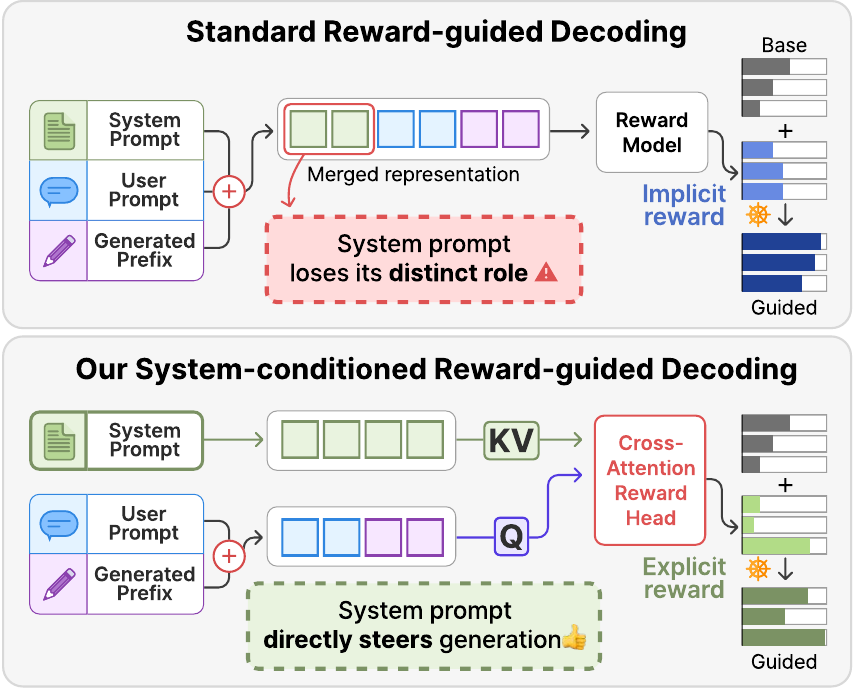}
\vspace{-0.15in}
\caption{
\textbf{
Comparison between standard and explicit system-conditioned reward-guided decoding.}
\textbf{(Top)} Standard reward-guided decoding merges the system prompt, user prompt, and generated prefix into a single representation, making the system prompt only an implicit reward signal.
\textbf{(Bottom)} Explicit system-conditioned reward-guided decoding preserves the system prompt as a distinct conditioning source. The cross-attention reward head uses system hidden states as key-value memory and user-prefix hidden states as the query to produce token-level rewards for steering generation.
}\label{fig:intro_concept}
\vspace{-0.1in}
\end{figure}

%% file: Figures/Figure2.tex
\begin{figure*}[!t]
\centering
\includegraphics[width=1.0\textwidth]{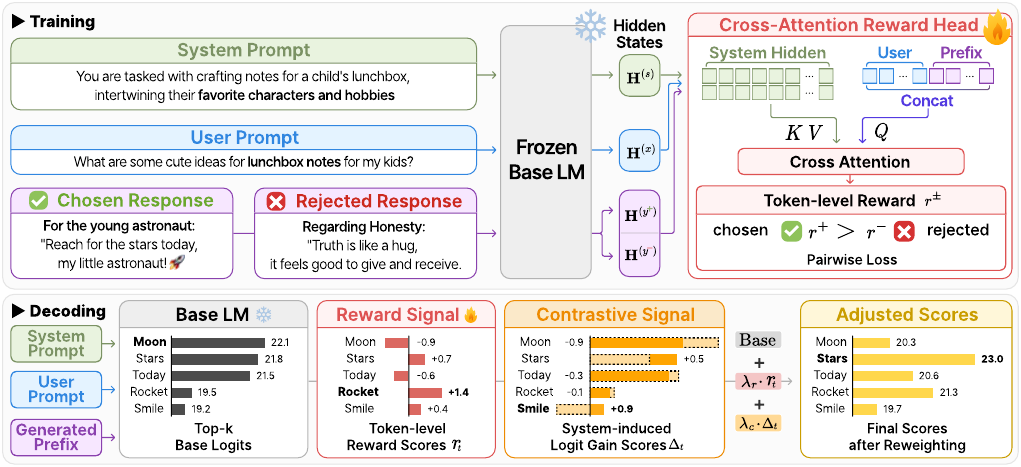}
\caption{
\textbf{An overview of \name{}.}
\textbf{Training stage:} a \textit{cross-attention reward head} attends from the user-query and response-prefix hidden states to the system prompt hidden states extracted from a frozen base LM, producing token-level rewards trained with a pairwise loss over chosen and rejected responses.
\textbf{Decoding stage:} top-$k$ candidates from the base LM are reweighted by combining the base logits with token-level reward scores and the contrastive signal, yielding final reranking scores better aligned with the system prompt.
}
\vspace{-0.2in}
\label{fig:method}
\end{figure*}

%% file: 2_related.tex
\section{Related Work}

\paragraph{System-prompt following.}
System prompts are increasingly used to specify multi-faceted behavioral requirements such as persona, style, format, and safety constraints \citep{lee2024aligning}.
Existing approaches for improving prompt adherence can be broadly categorized into \textit{training-time} and \textit{decoding-time} methods.
Training-time methods improve instruction following through instruction tuning and RLHF \citep{wei2021finetuned, ouyang2022training}, explicit priority learning across instruction sources \citep{wallace2024instruction}, or architectural separation of instruction segments \citep{wu2025instructional}.
While effective for constraints observed during training, these methods incur substantial retraining cost, may incur alignment tax, and are limited in adapting to unseen or diverse system prompt requirements.
Conversely, decoding-time methods keep the base LM frozen and intervene at inference, for instance by engineering prompts to emphasize instructions \citep{ye2024investigating}, refining outputs through self-feedback \citep{madaan2023self}, or contrasting token distributions induced by different prompt conditions \citep{shi2024trusting, zhu2025fly}.
However, these methods often rely on the implicit instruction-following ability of the base LM or treat the system prompt only as a distributional cue, rather than explicitly judging whether each candidate token satisfies the system prompt.

\paragraph{Reward-guided decoding.}
Reward-guided decoding steers a frozen base LM at inference by combining its next-token distribution with an auxiliary control signal.
Existing methods can be broadly categorized into \textit{classifier-based} and \textit{reward-based} approaches.
Classifier-based methods reweight LM logits using external discriminators, class-conditional language models, or expert and anti-expert pairs to control attributes such as toxicity, sentiment, or formality \citep{yang2021fudge, krause2021gedi, liu2021dexperts}.
While lightweight, these methods are typically designed for fixed attributes and do not naturally capture the layered and instance-specific constraints expressed in system prompts.
Meanwhile, reward-based methods use learned reward or critic models to score candidate continuations, either by applying sequence-level rewards to partial generations \citep{khanov2024args, deng2023reward} or by providing token-level reward signals for denser decoding-time guidance \citep{kim2023critic, xu2025genarm}.
However, prior work has focused on broad alignment or attribute-control objectives, leaving system-prompt following largely unexplored.
Moreover, reward models typically process the prompt and response as a single sequence, so the system prompt is absorbed into the prefix context.
\name{} addresses both issues by conditioning a learned reward on the system prompt and scoring token-level adherence to it directly.

%% file: 3_method.tex
\section{Method}\label{sec3}

\subsection{Overview and Problem Setup}\label{sec3.1}
\paragraph{Overview.} 
In this section, we propose \name{} (\textbf{Sy}stem-prompt following via \textbf{R}eward-g\textbf{u}ided \textbf{P}rediction), a decoding-time method for improving system-prompt adherence without updating the base LM.
The key idea is to make system-prompt following an explicit criterion during next-token selection.
Specifically, \name{} consists of two components that mirror the structure of this section.
First, a \textit{cross-attention reward head} (Section~\ref{sec3.2}) evaluates candidate next tokens by comparing the current generation state with the system prompt, which is treated as a separate key-value memory.
This allows the reward head to focus on the behavioral specification encoded in the system prompt while conditioning on the user query and generated prefix.
Second, a \textit{reward-guided decoding rule} (Section~\ref{sec3.3}) reranks the base LM's top-$k$ candidate tokens by combining base LM plausibility with both the learned reward score and the contrastive signal that measures how the system prompt shifts the base LM's own next-token distribution relative to a neutral prompt.
Figure~\ref{fig:method} illustrates the overall framework, and the procedure is summarized in Algorithms~\ref{alg:training} and~\ref{alg:decoding}.

\paragraph{Problem setup.}
We formulate the task of \textit{system-prompt-aware decoding} as follows: 
let $\mathcal{M}_\theta$ denote a base autoregressive LM with parameters $\theta$, and let $V$ denote its vocabulary.
At inference time, $\mathcal{M}_\theta$ is conditioned on a \textit{system prompt} $s$, which specifies the intended behavior of the model, such as persona, format, or safety constraints, together with a \textit{user input} $x$ that provides a concrete query under this specification. 
At each decoding step $t$, $\mathcal{M}_\theta$ produces next-token logits $\ell_t \in \mathbb{R}^{|V|}$ conditioned on $(s,x,y_{<t})$, from which conventional decoding algorithms select the next token $y_t$.
We denote the final-layer hidden states of a token sequence $u$ by $\mathbf{H}^{(u)} \in \mathbb{R}^{|u|\times d}$, where $d$ is the hidden size.
In particular, $\mathbf{H}^{(s)}$, $\mathbf{H}^{(x)}$, and $\mathbf{H}^{(y_{<t})}$ correspond to the system prompt, user input, and generated prefix, respectively.
The logits are then computed from the hidden state at the most recent position $\mathbf{h}_t$ (\textit{i.e.}, the last row of $\mathbf{H}^{(y_{<t})}$), as $\ell_t(v) = \mathbf{h}_t^\top \mathbf{e}_v$, where $\mathbf{e}_v$ denotes the frozen output-projection vector associated with token $v$

Then, our goal is to guide decoding while keeping $\mathcal{M}_\theta$ frozen, by intervening only in the token selection step at inference time.
To this end, we assign an auxiliary adherence score to candidate next tokens and combine it with the base LM logits, steering generation toward tokens that are both plausible under the base model and aligned with the system prompt $s$.

\subsection{Cross-attention Reward Head}\label{sec3.2}
Rather than merging the system prompt, user query, and generated prefix into a single hidden feature, we keep the system prompt as a separate memory that the reward head attends to at each decoding step. 
This design enables the proposed reward head to provide an explicit, token-level estimate of system-prompt adherence.

\paragraph{Cross-attention scoring.}
The proposed cross-attention reward head uses the system prompt hidden states $\mathbf{H}^{(s)}$ as a key-value memory.
The context representation at step $t$ is formed by concatenating the user input and generated prefix hidden states along the sequence dimension:
\begin{equation}
    \mathbf{H}_{\mathrm{ctx},t}
    =
    \operatorname{Concat}_{\mathrm{seq}}
    \left(
        \mathbf{H}^{(x)}, \mathbf{H}^{(y_{<t})}
    \right)
    \in \mathbb{R}^{L_t \times d},
\end{equation}
where $L_t = |x| + |y_{<t}|$.
Computing $\mathbf{H}^{(s)}$ separately allows the system prompt representation to serve as a stable memory that can be cached and reused across decoding steps.

With learnable matrices $W_Q, W_K, W_V \in \mathbb{R}^{d \times d}$ that constitute the trainable parameters of the cross-attention module, we compute queries, keys, and values as
$\mathbf{Q}_t = \mathbf{H}_{\mathrm{ctx},t}W_Q$,
$\mathbf{K}_s = \mathbf{H}^{(s)}W_K$, and
$\mathbf{V}_s = \mathbf{H}^{(s)}W_V$.
The cross-attention output is then
\begin{equation}
    \mathbf{A}_t
    =
    \mathrm{softmax}\!\left(
    \frac{\mathbf{Q}_t \mathbf{K}_s^\top}{\sqrt{d}}
    \right)
    \mathbf{V}_s,
    \label{eq:cross_attention}
\end{equation}
where padding positions in $\mathbf{H}^{(s)}$ are masked out before the softmax.
Each row of $\mathbf{A}_t$ summarizes the part of the system prompt that is most relevant to the corresponding context position.

We then fuse $\mathbf{H}_{\mathrm{ctx},t}$ and $\mathbf{A}_t$ by concatenating them along the feature dimension and passing the result through a two-layer MLP $g_\phi$ with a GELU activation:
\begin{equation}
    \mathbf{Z}_t
    =
    g_\phi\!\left(
    \operatorname{Concat}_{\mathrm{feat}}
    \left(
        \mathbf{H}_{\mathrm{ctx},t}, \mathbf{A}_t
    \right)
    \right)
    \in \mathbb{R}^{L_t \times d}.
\end{equation}
The representation at the most recent position is taken as the current decoding-state vector, $\mathbf{z}_t = (\mathbf{Z}_t)_{L_t}$, which plays the same role as $\mathbf{h}_t$ in the base LM.
For each candidate token $v \in V$, the reward head then assigns a token-level adherence score by
\begin{equation}
    r_t(v) = \mathbf{z}_t^\top \mathbf{e}_v .
    \label{eq:token_reward}
\end{equation}
This score estimates how well selecting token $v$ would continue the response in a direction consistent with the system prompt and the current generation state.

\paragraph{Pairwise training.}
System-prompt adherence is naturally judged at the response level, since a response must satisfy the behavioral specification as a whole.
We therefore train the reward head using system-prompt-conditioned preference pairs $(s,x,y^+,y^-)$, where $y^+$ better satisfies the system prompt than $y^-$.
For a response $y$, we obtain a sequence-level reward $R(s,x,y)$ by averaging the token-level rewards assigned to its generated tokens.
The reward head is then optimized with a Bradley--Terry objective \citep{bradley1952rank}:
\begin{equation}
    \mathcal{L}_{\mathrm{RM}}
    =
    -\mathbb{E}_{(s,x,y^+,y^-)}
    \log \sigma\!\left(\beta (R^+ - R^-)\right),
    \label{eq:rm_loss}
\end{equation}
where $R^\pm = R(s,x,y^\pm)$, $\sigma$ is the sigmoid function, and $\beta$ is a temperature parameter.
We remark that only the reward head parameters, namely the cross-attention projections $\{W_Q, W_K, W_V\}$ and the fusion MLP $g_\phi$, are updated during training; the base LM $\mathcal{M}_\theta$ remains frozen.

\subsection{Reward-guided Decoding}\label{sec3.3}
At inference time, our goal is to steer token selection without overriding the base LM's fluency.
We therefore restrict reranking to the top-$k$ candidates under the base LM logits.
This allows \name{} to intervene only among tokens that are already plausible under the base model, while avoiding full-vocabulary reward computation.

Formally, at each decoding step $t$, the cross-attention reward head produces the decoding-state vector $\mathbf{z}_t$, which yields a token-level reward $r_t(v)$ for each candidate $v$ via Eq.~\eqref{eq:token_reward}.
To complement the learned reward, we compute a contrastive signal that captures how the system prompt shifts the base LM's next-token logits relative to a neutral prompt:
\begin{equation}
    \Delta_t(v) = \ell_t(v) - \ell_t^{\mathrm{neu}}(v),
\end{equation}
where $\ell_t^{\mathrm{neu}}$ denotes the base LM logits obtained by replacing the system prompt with a neutral placeholder.
The final decoding rule combines the base LM logit, the learned reward, and the contrastive signal to rerank candidates:
\begin{equation}
    y_t
    =
    \arg\max_{v \in \mathcal{C}_t}
    \left[
        \ell_t(v)
        + \lambda_r r_t(v)
        + \lambda_c \Delta_t(v)
    \right],
    \label{eq:syrup_decoding}
\end{equation}
where $\lambda_r$ and $\lambda_c$ control the strengths of the learned reward signal and the contrastive signal, respectively.
Setting $\lambda_c=0$ yields decoding guided only by the learned reward, setting $\lambda_r=0$ yields decoding guided only by the contrastive signal, and setting both weights to zero recovers greedy decoding from the base LM.

The learned reward score $r_t$ and the contrastive signal $\Delta_t$ play complementary roles.
The reward score provides a trained token-level estimate of system-prompt adherence through the cross-attention reward head.
In contrast, $\Delta_t$ captures how much the system prompt shifts the base LM's own next-token preference relative to a neutral prompt.
By combining these two signals, \name{} uses both an explicit adherence evaluator and the base LM's intrinsic sensitivity to the system prompt.

%% file: 4_experiment.tex
\section{Experiments}
\input{Tables/Table_main_mf}
\subsection{Setups}
\paragraph{Datasets and metrics.}
For training the auxiliary reward models used by the reward-based baselines and the reward head used by our method, we use the \textit{Multifaceted-Collection-DPO} dataset~\citep{lee2024aligning}, which contains approximately 65K training examples with system prompts, user queries, and preference pairs of chosen and rejected responses.
We provide further details on all datasets, including dataset specifications and statistics, in Appendix~\ref{app:datasets}; Table~\ref{tab:dpo_example} shows a representative training example.
Then, to evaluate whether \name{} improves system-prompt following, we consider three complementary evaluation settings:

\noindent(1) \textit{Multifaceted-Bench.}
First, for evaluating system-prompt following under diverse user values, we use Multifaceted-Bench \citep{lee2024aligning}.
It augments existing instruction-following benchmarks with unseen system messages that specify target behaviors or preferences.
We evaluate responses on five Multifaceted-Bench tasks, corresponding to the multifaceted versions of AlpacaEval, Koala, MT-Bench, Self-Instruct, and FLASK.
Following the benchmark protocol, we use instance-specific LLM-as-a-Judge rubrics tailored to each system message and reference answer.
We use \texttt{gpt-4o-mini} \citep{openai2024gpt4omini} as the evaluator LLM, repeat evaluation three times for each response, and report the averaged score for each task and across all tasks.

\noindent(2) \textit{IFEval.}
Next, for out-of-distribution (OOD) instruction following evaluation under verifiable constraints, we use IFEval \citep{zhou2023instruction}.
IFEval programmatically checks whether a response satisfies explicitly specified instructions, complementing the LLM-based evaluation in Multifaceted-Bench.
Following \citet{venkateswaran2026spotlight}, we use the version where each prompt is rewritten with \texttt{gpt-4o-mini} to separate task content from instruction constraints.

\noindent(3) \textit{IHEval.}
Finally, we use the Rule Following subset of IHEval~\citep{zhang2025iheval} to evaluate system-level instruction priority in single- and multi-turn settings with aligned or conflicting system and user instructions.
For both IFEval and IHEval, \name{} reuses the corresponding reward head trained on Multifaceted-Collection-DPO and the Multifaceted-Bench decoding configuration without benchmark-specific training or tuning.
For consistency, we report strict prompt-level accuracy \citep{liu2024deepseek, yang2025qwen3} on a 0--100 scale for both benchmarks, with all four IHEval metrics provided in the Appendix.

\paragraph{Baselines.}
We compare \name{} against several baselines.
(1) \textit{Base} generates responses using the vanilla LLM without any additional prompting strategy, reward model, or decoding-time steering.
For \textbf{prompting baselines} (2--4), we include methods that modify the input format while keeping the base model and decoding procedure unchanged.
These include
(2) \textit{*-marked} and
(3) \textit{"-marked}, which enclose the system prompt with explicit symbols to make the instruction boundary more salient,
and (4) \textit{Few-shot}, which prepends exemplars to guide the model toward the desired response behavior.
For \textbf{decoding-time baselines} (5--9), we include methods that modify generation at inference time using contrastive, reward-based, or search-based signals.
Specifically, we evaluate
(5) \textit{OPAD} \citep{zhu2025fly}, which performs principle-guided decoding by exploiting the discrepancy between constrained and unconstrained policies,
(6) \textit{GenARM} \citep{xu2025genarm}, which steers generation through model arithmetic with a trained reward surrogate,
(7) \textit{ARGS} \citep{khanov2024args}, which reranks candidate tokens using a reward model during decoding,
(8) \textit{Bo$N$} \citep{gui2024bonbon}, which samples multiple candidate responses and selects the highest-scoring one according to a reward model,
and (9) \textit{DeAL} \citep{huang2025deal}, which uses reward-guided lookahead search.
{All auxiliary reward models are trained on the same 65K Multifaceted-Collection-DPO dataset using comparable parameter-efficient training protocols, rather than using off-the-shelf reward models.}

\paragraph{Implementation details.}
We conduct experiments with three backbone LLMs: \texttt{Qwen2.5-1.5B-Instruct} \citep{qwen2025qwen25technicalreport}, \texttt{Llama-3.2-3B-Instruct}, and \texttt{Llama-3.1-8B-Instruct} \citep{grattafiori2024llama}.
For each backbone, we train a model-specific reward head on Multifaceted-Collection-DPO, while keeping the backbone LLM frozen.
We use AdamW \citep{loshchilov2017decoupled} with a learning rate of $1 \times 10^{-4}$, a weight decay of 0.01, and a linear learning-rate scheduler with a warmup ratio of 0.1.
The reward head is trained for 2 epochs with a pairwise preference objective in Eq.~\ref{eq:rm_loss}, which assigns higher rewards to chosen responses than rejected responses, using a Bradley--Terry temperature of $\beta=1.0$.
At inference time, \name{} guides next-token prediction {using the learned reward and contrastive signals} according to Eq.~\ref{eq:syrup_decoding}.
The reward head attends to the system-prompt hidden states and scores candidate next tokens conditioned on the user prompt and generated prefix.
Unless otherwise specified, we rerank the top-$k$ candidate tokens using $k=10$, $\lambda_r=0.05$, and $\lambda_c=2.0$.
We select these hyperparameters on a fixed Multifaceted-Bench development split (Appendix~\ref{app:dev_selection}).
We refer to configurations using only the learned reward or contrastive signal as \textit{Reward-only} and \textit{CS-only}, respectively, while \name{} denotes their combination.
For IFEval and IHEval, we reuse the reward head and decoding hyperparameters used on \textit{Multifaceted-Bench}, without additional training or task-specific tuning.
All methods are evaluated with the same maximum generation length of 1024 tokens and deterministic greedy decoding, except for sampling-based baselines such as \textit{Bo$N$}.

\subsection{Main Results}
Table~\ref{tab:main_mf} presents the in-domain evaluation results on Multifaceted-Bench. The reward head is trained exclusively on Multifaceted-Collection-DPO, while Multifaceted-Bench is used only for evaluation.
\name{} achieves the best score on all 15 model-task combinations, improving the average over the strongest baseline by 5.58\%, 3.67\%, and 3.47\% on \texttt{Qwen2.5-1.5B}, \texttt{Llama-3.2-3B}, and \texttt{Llama-3.1-8B}, respectively.
This demonstrates consistent gains across both model families and evaluation subsets.

Comparing baseline families, prompting methods yield only modest improvements, while decoding-time methods generally perform better.
On \texttt{Qwen2.5-1.5B}, the strongest prompting and decoding-time baselines reach 3.56 and 3.74, respectively, while GenARM is the strongest competing method on the Llama backbones (4.08 and 4.18).
The advantage of \name{} is pronounced on the smaller \texttt{Qwen2.5-1.5B} model, where it improves the average score by 21.54\% over vanilla decoding (from 3.25 to 3.95).
This suggests that token-level guidance using system-prompt-aware reward and contrastive signals can substantially improve adherence, particularly when the base model has limited instruction-following capacity.
Additional scaling results on \texttt{Qwen3-14B} are reported in Appendix~\ref{app:qwen14}.

\input{Tables/Table_main_ifeval}
Table~\ref{tab:main_ifeval} examines OOD transfer on IFEval, where the full \name{} configuration is applied to verifiable instruction-following constraints without IFEval-specific training or tuning.
Despite this distribution shift, \name{} achieves the highest numerical average strict prompt-level accuracy across the three backbones, improving the mean over the strongest baseline, GenARM, from 58.50 to 59.18.
It achieves the best performance on \texttt{Qwen2.5-1.5B} and ranks second-best on both \texttt{Llama} backbones, while consistently outperforming vanilla decoding.
These results indicate that \name{} transfers from preference-oriented system-prompt following to verifiable instruction adherence.

\input{Tables/Table_main_iheval}
Table~\ref{tab:iheval_main} extends OOD evaluation to instruction-hierarchy following under aligned and conflicting instructions.
\name{} achieves the best performance across all four settings, improving the average strict prompt-level accuracy over GenARM from 26.90 to 35.90.
The improvement is pronounced in the multi-turn conflict setting, where accuracy increases from 9.61 to 24.95.
This suggests that \name{} more reliably preserves system-level instruction priority when user instructions conflict with the system prompt across turns.
Together, the IFEval and IHEval results demonstrate effective transfer to both verifiable constraints and hierarchical instruction following without benchmark-specific training or tuning.

We additionally evaluate the \texttt{Qwen2.5-1.5B} backbone on RealGuardrails \citep{mu2025closer}, and the Chinese SysBench \citep{qin2024sysbench} without target-specific tuning, with the results reported in Appendix~\ref{app:broader_ood}.

\subsection{Analyses}
Unless otherwise specified, all analyses in this section use \texttt{Qwen2.5-1.5B-Instruct} on Multifaceted-Bench using top-$k$ reranking with $k=10$, $\lambda_r=0.05$, and $\lambda_c=2.0$.
This representative setting follows the main configuration and lets us examine \name{} without additional analysis-specific tuning.

\paragraph{Inference efficiency.}
\input{Tables/Table_latency}
Table~\ref{tab:latency} compares the inference cost of \name{} with prompting and decoding-time baselines.\footnote{
We exclude Bo$N$ because it requires the model to generate multiple responses and then selects among them, making its cost primarily depend on the number of sampled candidates rather than token-level guidance.}
We report wall-clock latency per final output token, peak GPU memory, and relative latency normalized to the Base method.
The learned reward component of \name{} uses a lightweight token-level reward head rather than a full trajectory-level reward model.
Reward-only increases latency by only $1.04\times$, while CS-only and the full \name{} configuration incur $2.02\times$ and $2.18\times$ latency, respectively.
The additional cost of the latter two configurations primarily comes from the extra forward pass required to compute the contrastive signal.

Nevertheless, this cost remains moderate compared with more expensive decoding-time baselines.
GenARM requires $2.80\times$ the Base latency, while ARGS and DeAL require $5.13\times$ and $16.83\times$, respectively.
Moreover, \name{} keeps peak GPU memory close to the Base model, using only 3.1 GB compared with 6.4 GB for GenARM and over 10 GB for ARGS and DeAL.
Unlike these baselines, \name{} does not introduce a separate LLM for reward modeling.
Its learned component is a lightweight token-level reward head, while the contrastive signal reuses the frozen base LM.
Prompting-only methods introduce almost no additional cost because they preserve the original decoding procedure, but their performance gains are limited as shown in Table~\ref{tab:main_mf}.
Overall, \name{} provides a favorable trade-off between inference-time cost, memory usage, and system-prompt adherence.

\paragraph{Role of reward and contrastive signals.}
\input{Figures/Figure3}
Figure~\ref{fig:weight_effect} shows the effect of varying the reward weight $\lambda_r$ and the contrastive weight $\lambda_c$.
Reward-only and CS-only achieve average scores of 3.43 and 3.68, respectively, while their combination reaches 3.95.
Thus, CS is the stronger standalone signal on Multifaceted-Bench, while the learned reward provides complementary guidance in the full configuration.
For reward guidance, performance peaks at a moderate value of $\lambda_r=0.05$, while a large reward weight degrades the score.
This suggests that the learned reward should guide token selection without dominating the base LM.
For the contrastive signal, performance improves as $\lambda_c$ increases up to $\lambda_c=2.0$, then slightly drops at $\lambda_c=3.0$.
This indicates that the contrastive signal provides a strong system-prompt alignment cue, but excessive weighting can over-steer generation.
The starred points indicate the default values used in our main experiments.

\paragraph{Token-level signal interaction.}
\input{Figures/Figure4}
Figure~\ref{fig:hybrid_token_source} examines how the learned reward and contrastive signals interact during token selection.
Overall, 9.8\% of \name{}-selected tokens are \textit{\name{}-unique}, meaning that they are not top-ranked by the base LM, reward signal, or contrastive signal alone.
The combined score can therefore select tokens that arise specifically from integrating the two guidance signals rather than copying the top choice of any individual signal.
Additional rank- and correlation-based diagnostics are reported in Appendix~\ref{app:token_diagnostics}.

\paragraph{Effect of system-prompt conditioning.}
\input{Figures/Figure5}
Figure~\ref{fig:reward_head} evaluates whether explicit system-prompt conditioning is necessary for the reward head.
We compare three reward-head variants under the Reward-only and full \name{} configurations.
\textit{No-system} removes direct system-prompt access from the reward head, so token scores are computed only from the user prompt and generated prefix.
\textit{Single-stream} keeps the system prompt in the input sequence, but encodes the system prompt, user prompt, and generated prefix as a single sequence and uses the last-token representation, relying on implicit propagation through self-attention.
In contrast, \textit{Cross-attention}, used in \name{}, treats the system prompt as a separate memory and directly attends to its hidden states when producing token-level rewards.
The separation among architectures is clearest under Reward-only, where the reward head is the sole steering signal.
Their narrower gaps under the full \name{} configuration indicate that the shared contrastive signal narrows the differences among reward-head variants.

\paragraph{Dimension-wise performance.}
\input{Tables/Table_dimension_main}
Table~\ref{tab:dimension_main} compares performance across the four evaluation dimensions defined in Multifaceted-Bench.
\name{} achieves the highest score in every dimension, improving over the strongest baseline in each dimension by 4.77\%, 5.29\%, 6.69\%, and 5.47\% in style, background knowledge, informativeness, and harmlessness, respectively.
These consistent improvements indicate that the gains are not confined to stylistic adherence, with the largest relative improvement appearing in informativeness.
Results for all baselines are provided in Appendix~\ref{app:dimension_wise}.

\paragraph{Evaluation robustness.}
\input{Tables/Table_judge_main}
Table~\ref{tab:judge_main} reports full-set Multifaceted-Bench averages under \texttt{gpt-4o-mini} and \texttt{DeepSeek-V3} \citep{liu2024deepseek} using the same rubrics.
The method ordering remains unchanged across the two evaluators, with \name{} achieving the highest average under both.
This consistency indicates that the in-domain advantage is not specific to \texttt{gpt-4o-mini}.
Full per-subset results and evaluation details are in Appendix~\ref{app:judge_robustness}.

%% file: Tables/Table_main_mf.tex
\begin{table*}[t]
\caption{\textbf{Main results on Multifaceted-Bench.} 
LLM-judge scores on a 1--5 scale across five evaluation subsets and their average using three base models. 
Alp., Koa., MT, SI, and FL. denote mf-AlpacaEval, mf-Koala, mf-MT-Bench, mf-Self-Instruct, and mf-FLASK, respectively.
\textbf{Bold} indicates the best performance, and \underline{underlined} values indicate the second-best performance within each model.}
\label{tab:main_mf}
\centering

\fontsize{7.5pt}{8pt}\selectfont
\setlength{\tabcolsep}{1.8pt}
\renewcommand{\arraystretch}{1.08}

\resizebox{\textwidth}{!}{%
\begin{tabular}{l|ccccc|c|ccccc|c|ccccc|c}
\toprule[1.4pt]

& \multicolumn{6}{c|}{\texttt{Qwen2.5-1.5B}}
& \multicolumn{6}{c|}{\texttt{Llama-3.2-3B}}
& \multicolumn{6}{c}{\texttt{Llama-3.1-8B}} \\
\cmidrule(lr){2-7} \cmidrule(lr){8-13} \cmidrule(lr){14-19}

\textbf{Method}
& Alp. & Koa. & MT & SI & FL. & Avg.
& Alp. & Koa. & MT & SI & FL. & Avg.
& Alp. & Koa. & MT & SI & FL. & Avg. \\

\midrule[1pt]

Base
& 3.36 & 3.23 & 3.53 & 3.12 & 2.99 & 3.25
& 3.86 & 3.94 & 3.90 & 3.78 & 3.55 & 3.81
& 3.98 & 3.94 & 4.05 & 3.92 & 3.67 & 3.91 \\

\midrule

*-marked
& 3.61 & 3.80 & 3.69 & 3.41 & 3.30 & 3.56
& 3.88 & 3.98 & 3.90 & 3.83 & 3.60 & 3.84
& 4.08 & 4.02 & 4.05 & 3.93 & 3.74 & 3.96 \\

"-marked
& 3.50 & 3.70 & 3.62 & 3.28 & 3.14 & 3.45
& 3.73 & 3.87 & 3.88 & 3.64 & 3.35 & 3.69
& 3.88 & 4.00 & 4.00 & 3.82 & 3.59 & 3.86 \\

Few-shot
& 3.42 & 3.56 & 3.53 & 3.29 & 2.95 & 3.35
& 3.94 & 3.90 & 3.88 & 3.72 & 3.40 & 3.77
& 4.09 & 4.04 & 4.03 & 3.92 & 3.68 & 3.95 \\

\midrule

OPAD
& 3.63 & 3.71 & 3.61 & 3.34 & 3.23 & 3.51
& 3.98 & 4.11 & 4.05 & 3.90 & 3.73 & 3.95
& 4.19 & 4.24 & 4.16 & 4.06 & 3.92 & 4.11 \\

GenARM
& \underline{3.86} & 3.74 & \underline{3.83} & \underline{3.67} & \underline{3.60} & \underline{3.74}
& \underline{4.30} & \underline{4.26} & \underline{4.09} & \underline{3.97} & \underline{3.80} & \underline{4.08}
& \underline{4.30} & \underline{4.28} & \underline{4.21} & \underline{4.12} & \underline{3.98} & \underline{4.18} \\

ARGS
& 3.25 & 3.19 & 3.41 & 3.05 & 3.05 & 3.19
& 3.87 & 3.99 & 4.07 & 3.81 & 3.54 & 3.86
& 4.07 & 4.07 & 4.05 & 3.98 & 3.80 & 4.00 \\

Bo$N$
& 3.73 & \underline{3.83} & 3.70 & 3.35 & 3.37 & 3.60
& 4.00 & 4.08 & 4.01 & 3.83 & 3.65 & 3.92
& 4.12 & 4.12 & 4.15 & 3.97 & 3.87 & 4.04 \\

DeAL
& 3.51 & 3.45 & 3.52 & 3.18 & 3.19 & 3.37
& 3.94 & 4.03 & 3.99 & 3.83 & 3.61 & 3.88
& 4.13 & 4.08 & 4.14 & 3.99 & 3.78 & 4.02 \\

\midrule

\rowcolor[HTML]{D9D9D9}
\name{}
& \textbf{4.08} & \textbf{4.07} & \textbf{3.98} & \textbf{3.97} & \textbf{3.66} & \textbf{3.95}
& \textbf{4.36} & \textbf{4.34} & \textbf{4.27} & \textbf{4.13} & \textbf{4.06} & \textbf{4.23}
& \textbf{4.44} & \textbf{4.39} & \textbf{4.26} & \textbf{4.36} & \textbf{4.17} & \textbf{4.32} \\

\bottomrule[1.4pt]
\end{tabular}%
}

\vspace{-0.05in}
\end{table*}

%% file: Tables/Table_main_ifeval.tex
\begin{table}[t]
    \centering
    \caption{\textbf{Main results on IFEval.}
    Strict prompt-level accuracy on a 0--100 scale.
    \textbf{Bold} indicates the best performance, and \underline{underlined} values indicate the second-best performance within each model.}
    \label{tab:main_ifeval}
    
    \resizebox{\columnwidth}{!}{%
        \begin{tabular}{l|ccc}
            \toprule[1.5pt]
            \textbf{Method} 
            & \texttt{Qwen2.5-1.5B} 
            & \texttt{Llama-3.2-3B} 
            & \texttt{Llama-3.1-8B} \\
            \midrule[1pt]
            Base      & 31.54 & 62.34 & 65.31 \\
            \midrule[0.3pt]
            *-marked     & 28.39 & 62.34 & 66.60 \\
            "-marked     & 29.50 & 60.11 & 63.27 \\
            Few-shot     & 31.54 & 48.79 & 65.12 \\
            \midrule[0.3pt]
            OPAD         & 37.66 & \underline{65.12} & 69.20 \\
            GenARM       & \underline{41.00} & 63.08 & \textbf{71.43} \\
            ARGS         & 35.25 & 61.41 & 69.20 \\
            Bo$N$          & 38.03 & 64.94 & 68.09 \\
            DeAL         & 36.92 & \textbf{67.53} & 68.10 \\
            \midrule[0.3pt]
            \rowcolor[HTML]{D9D9D9}
            \name{} 
                         & \textbf{41.19} & \underline{65.12} & \underline{71.24} \\
            \bottomrule[1.5pt]
        \end{tabular}%
    }
\vspace{-0.15in}
\end{table}

%% file: Tables/Table_main_iheval.tex
\begin{table}[t]
    \centering
    \caption{\textbf{Main results on IHEval Rule Following.}
    Strict prompt-level accuracy on a 0--100 scale for \texttt{Qwen2.5-1.5B-Instruct}.}
    \label{tab:iheval_main}
    \small
    \setlength{\tabcolsep}{4pt}

    \begin{tabular}{l|rr|rr}
        \toprule[1.4pt]
        \multirow{2}{*}{\textbf{Method}}
        & \multicolumn{2}{c|}{\textbf{Single-turn}}
        & \multicolumn{2}{c}{\textbf{Multi-turn}} \\
        \cmidrule(lr){2-3}\cmidrule(lr){4-5}
        & \textbf{Aligned}
        & \textbf{Conflict}
        & \textbf{Aligned}
        & \textbf{Conflict} \\
        \midrule[1pt]

        Base
        & 28.65 & 12.20 & 23.66 & 6.47 \\

        *-marked
        & 25.88 & 10.54 & 22.74 & 4.99 \\

        OPAD
        & 36.97 & 15.90 & 34.38 & 6.10 \\

        GenARM
        & \underline{40.67}
        & \underline{19.22}
        & \underline{38.08}
        & \underline{9.61} \\

        \rowcolor[HTML]{D9D9D9}
        \name{}
        & \textbf{45.10}
        & \textbf{24.21}
        & \textbf{49.35}
        & \textbf{24.95} \\

        \bottomrule[1.4pt]
    \end{tabular}
\vspace{-0.15in}
\end{table}

%% file: Tables/Table_latency.tex
\begin{table}[t]
    \centering
    \caption{
    \textbf{Inference cost comparison.}
    Inference efficiency measured by latency per final output token, peak GPU memory, and relative latency normalized to the Base method.
    }
    \label{tab:latency}

    \resizebox{\columnwidth}{!}{%
        \begin{tabular}{l|ccc}
            \toprule[1.5pt]
            \textbf{Method}
            & \textbf{ms/token} 
            & \textbf{Peak mem. (GB)} 
            & \textbf{Latency} \\
            \midrule[1pt]
            Base        & 24.4  & 2.9  & $1.00\times$ \\
            \midrule[0.3pt]
            *-marked    & 24.3  & 2.9  & $1.00\times$ \\
            "-marked    & 24.0  & 2.9  & $0.98\times$ \\
            Few-shot    & 24.8  & 3.1  & $1.02\times$ \\
            \midrule[0.3pt]
            OPAD        & 48.7  & 3.0  & $2.00\times$ \\
            GenARM      & 68.2  & 6.4  & $2.80\times$ \\
            ARGS        & 125.0 & 10.2 & $5.13\times$ \\
            DeAL        & 410.3 & 10.4 & $16.83\times$ \\
            \midrule[0.3pt]
            Reward-only & 25.5  & 3.1  & $1.04\times$ \\
            CS-only     & 49.1  & 3.0  & $2.02\times$ \\
            \rowcolor[HTML]{D9D9D9}
            \name{}     & 53.0  & 3.1  & $2.18\times$ \\
            \bottomrule[1.5pt]
        \end{tabular}%
    }
    \vspace{-0.15in}
\end{table}

%% file: Figures/Figure3.tex
\begin{figure}[!t]
    \centering
    \includegraphics[width=\linewidth]{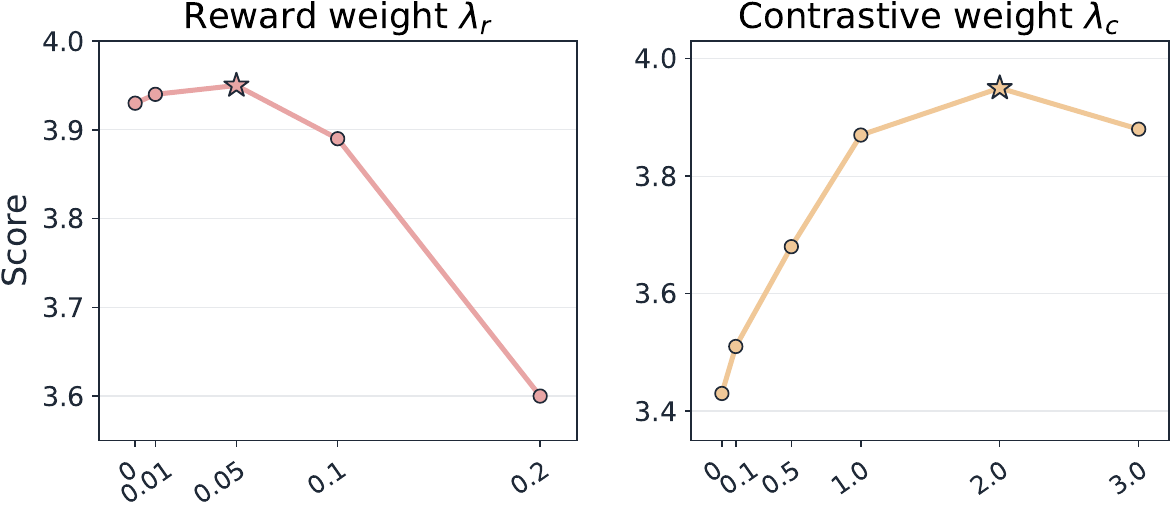}
    \caption{
    \textbf{Effect of reward and contrastive weights.}
    Average LLM-judge scores when varying the reward weight $\lambda_r$ and the contrastive weight $\lambda_c$.
    Stars indicate the default values used in our main experiments.
    }
    \label{fig:weight_effect}
\end{figure}

%% file: Figures/Figure4.tex
\begin{figure}[!t]
    \centering
    \includegraphics[width=\linewidth]{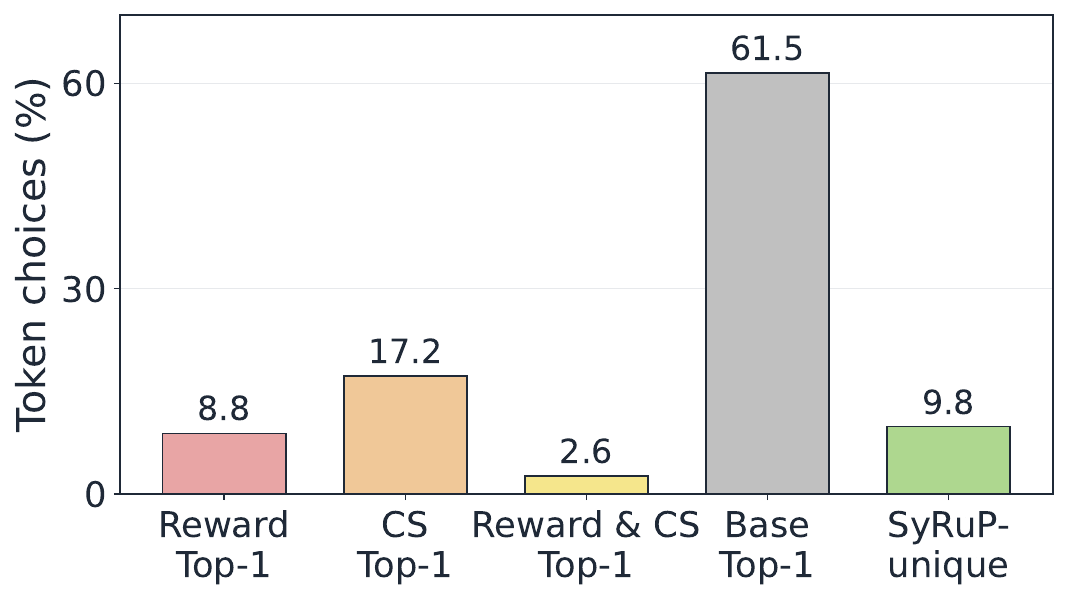}
    \caption{
    \textbf{Breakdown of token choices.}
    Each segment shows the share of \name{}-selected tokens that match the Top-1 token under the base LM, reward signal, contrastive signal (CS), or both.
    \textit{\name{}-unique} indicates tokens that are not Top-1 under any individual signal, capturing choices that arise only from \name{}.
    }
    \label{fig:hybrid_token_source}
\vspace{-0.15in}
\end{figure}

%% file: Figures/Figure5.tex
\begin{figure}[t] 
    \centering 
    \includegraphics[width=\linewidth]{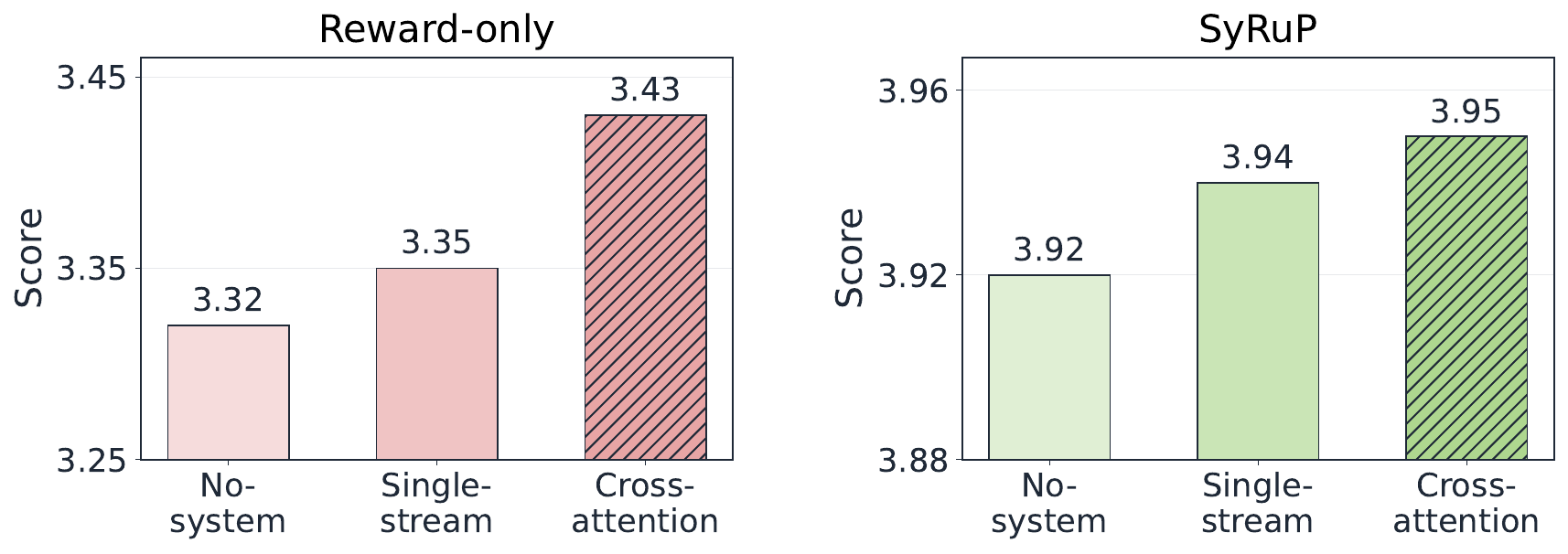} 
    \caption{\textbf{Effect of reward-head conditioning.} 
    Average LLM-judge scores on Multifaceted-Bench for three reward-head variants.}
    \label{fig:reward_head} 
    \vspace{-0.1in} 
\end{figure}

%% file: Tables/Table_dimension_main.tex
\begin{table}[t]
    \centering
    \caption{\textbf{Dimension-wise results on Multifaceted-Bench.}
    LLM-judge scores across four evaluation dimensions.
    Back., Info., and Harm. denote background knowledge, informativeness, and harmlessness, respectively. \textbf{Bold} and \underline{underlined} values indicate the best and second-best results, respectively.}    
    \label{tab:dimension_main}
    \small
    \renewcommand{\arraystretch}{1.05}

    \begin{tabular}{l|cccc}
        \toprule[1.4pt]
        \textbf{Method}
        & \textbf{Style}
        & \textbf{Back.}
        & \textbf{Info.}
        & \textbf{Harm.} \\
        \midrule[1pt]

        Base
        & 3.24 & 3.32 & 3.06 & 3.37 \\

        *-marked
        & 3.56 & 3.63 & 3.36 & 3.69 \\

        OPAD
        & 3.50 & 3.55 & 3.35 & 3.62 \\

        GenARM
        & \underline{3.77}
        & \underline{3.78}
        & \underline{3.59}
        & \underline{3.84} \\

        \rowcolor[HTML]{D9D9D9}
        \name{}
        & \textbf{3.95}
        & \textbf{3.98}
        & \textbf{3.83}
        & \textbf{4.05} \\

        \bottomrule[1.4pt]
    \end{tabular}
\end{table}

%% file: Tables/Table_judge_main.tex
\begin{table}[t]
    \centering
    \caption{\textbf{Evaluator robustness.}
    LLM-judge scores on a 1--5 scale, averaged across the five Multifaceted-Bench evaluation subsets, using two evaluators with the same rubrics.}
    \label{tab:judge_main}
    \small
    \setlength{\tabcolsep}{11pt}
    \renewcommand{\arraystretch}{1.08}

    \begin{tabular}{l|cc}
        \toprule[1.4pt]
        \textbf{Method}
        & \texttt{gpt-4o-mini}
        & \texttt{DeepSeek-V3} \\
        \midrule[1pt]

        Base
        & 3.25
        & 3.38 \\

        *-marked
        & 3.56
        & 3.69 \\

        OPAD
        & 3.51
        & 3.62 \\

        GenARM
        & \underline{3.74}
        & \underline{3.86} \\

        \rowcolor[HTML]{D9D9D9}
        \name{}
        & \textbf{3.95}
        & \textbf{4.04} \\

        \bottomrule[1.4pt]
    \end{tabular}
    \vspace{-0.1in}
\end{table}

%% file: 5_conclusion.tex
\section{Conclusion}
We propose \name{}, a decoding-time framework for improving system-prompt adherence without updating the base LM.
\name{} combines a cross-attention reward head that treats the system prompt as a separate conditioning source with a training-free contrastive signal derived from the base LM.
Together, these signals provide auxiliary token-level guidance for reranking plausible next-token candidates throughout generation.
Extensive experiments on system-prompt benchmarks show that \name{} improves adherence to diverse system prompts.
Collectively, our findings highlight the importance of explicitly modeling system-prompt conditioning during decoding and suggest a practical direction for controllable LLM generation under increasingly complex prompt specifications.

\newpage
\section*{Limitations}
While \name{} improves system-prompt adherence without updating the base LM, it has several limitations that suggest directions for future work.
First, the hybrid decoding strategy introduces additional inference overhead.
Although the reward head itself is lightweight, the contrastive signal requires an extra forward pass during decoding.
This makes the full hybrid variant slower than standard decoding, especially for long outputs.
Future work could reduce this cost by activating hybrid guidance only when needed, such as at uncertain decoding steps or at positions that are likely to be relevant to system-prompt following \citep{wang2025adacad}.
Other promising directions include early-exit strategies that avoid redundant computation when the guided and unguided distributions are sufficiently similar \citep{schuster2022confident}.

In addition, directly incorporating reward scores into token selection may lead to oversteering in some cases~\citep{gao2022scalinglawsrewardmodel, rafailov2024scaling}.
Although the top-$k$ constraint reduces this risk, the model may still over-prefer high-reward tokens at the expense of fluency, task relevance, or naturalness.
Future work should study calibration methods, adaptive reward weights, reward uncertainty estimation, and safety filtering to keep reward guidance helpful without dominating the base model's generation behavior.

\section*{Broader Impact and Ethical Implications}
The deployment of \name{} can improve the reliability of LLM-based applications by helping models follow developer-specified roles, output formats, domain guidelines, and safety policies more consistently without updating the base LM.
This may provide a flexible and resource-efficient way to strengthen system-level control across different deployment settings.
However, stronger system-prompt adherence may also make models follow hidden, undesirable, or overly restrictive system-level objectives more consistently, especially when users are not aware of the instructions or reward signals guiding generation.
In addition, reward-guided decoding may amplify biases present in the training data or cause oversteering if the reward head is poorly calibrated.
Future systems should therefore incorporate transparency, auditing, bias evaluation, and safety-aware monitoring to ensure that guided behavior remains reliable, fair, and aligned with responsible deployment practices.

\section*{Acknowledgments}
All authors are affiliated with the Department of Artificial Intelligence at Yonsei University.
This research was supported in part by the Institute for Information \& Communications Technology Planning \& Evaluation (IITP) grant funded by the Korea government (MSIT) (No. RS-2020-II201361, Artificial Intelligence Graduate School Program (Yonsei University); No. RS-2025-02215344, Development of AI Technology with Robust and Flexible Resilience Against Risk Factors).

%% file: 6_appendix.tex
\newpage

\section{Datasets}
\label{app:datasets}
\subsection{Dataset specification}

\paragraph{Multifaceted-Collection-DPO.}
We use Multifaceted-Collection-DPO~\citep{lee2024aligning} as the training dataset for our reward head.
The dataset is designed for aligning language models to diverse and fine-grained user preferences expressed through system messages.
The instructions are collected from several existing preference datasets, including Nectar \citep{zhu2024starlingb} and OpenHermesPreferences \citep{open_hermes_preferences}.
Each system message describes an individual's multifaceted preference along four high-level dimensions: style, background knowledge, informativeness, and harmlessness.
Each example consists of a system message, a user instruction, a chosen response, and a rejected response.
The chosen response is generated to follow the given system message and instruction, whereas the rejected response follows a different system message for the same instruction.
Thus, the rejected response is not necessarily low-quality or irrelevant; rather, it is plausible but less aligned with the target system-level preference.
For the same dialogue context, this paired structure provides direct supervision for learning whether a candidate continuation is aligned with the intended system-level preference.
Table~\ref{tab:dpo_example} shows a representative example from Multifaceted-Collection-DPO.

\paragraph{Multifaceted-Bench.}
We use Multifaceted-Bench~\citep{lee2024aligning} as the main in-domain evaluation benchmark.
It evaluates whether a model can generate responses that are tailored to explicit, context-specific preferences described in system messages.
The benchmark contains instructions sampled from five existing evaluation sources, including AlpacaEval 2.0 \citep{dubois2024length}, FLASK \citep{ye2024flask}, Koala \cite{koala_blogpost_2023}, MT-Bench \citep{zheng2023judging}, and Self-Instruct \citep{wang-etal-2023-self-instruct}.
These sources cover diverse instruction-following scenarios, such as open-ended helpfulness evaluation, multi-turn dialogue, knowledge-intensive question answering, reasoning, and general user-oriented tasks.
For each instruction, the benchmark constructs multiple preference-conditioned instances by pairing the instruction with a synthetic system message and a reference answer.
Each instance also provides customized 1--5 score rubrics for the four preference dimensions: style, background knowledge, informativeness, and harmlessness.
This benchmark is therefore well aligned with our goal of improving system-prompt adherence beyond generic instruction following.

\paragraph{IFEval.}
We additionally evaluate on IFEval~\citep{zhou2023instruction}, an out-of-domain benchmark for instruction-following evaluation that focuses on verifiable instructions such as length constraints, keyword requirements, formatting constraints, and punctuation restrictions. Unlike Multifaceted-Bench, which evaluates preference-conditioned response quality, IFEval measures whether a model satisfies explicitly checkable constraints, allowing us to assess whether our method improves general instruction adherence without relying on the preference structure used during training.

Since the original IFEval prompts interleave the primary task with its associated instructions, we follow the preprocessing procedure from \citet{venkateswaran2026spotlight} and rewrite the prompts using the \texttt{gpt-4o-mini} API~\citep{openai2024gpt4omini} to disentangle the two components, guided by the template shown in Figure~\ref{fig:rewrite-prompt}. Table~\ref{tab:ifeval-rewrite-examples} illustrates the resulting structural changes. After rewriting, we place the separated instructions in the system prompt and the task in the user message, matching the input format used in Multifaceted-Bench.

\paragraph{IHEval.}
We use the Rule Following subset of IHEval~\citep{zhang2025iheval} to evaluate whether models prioritize system-level instructions within instruction hierarchies.
It includes single-turn and multi-turn settings in which system and user instructions are either aligned or conflicting.
The benchmark reports strict and loose accuracy at both the prompt and instruction levels.

\input{Tables/appendix/Table_ifeval1}
\input{Tables/appendix/Table_ifeval2}

\subsection{Dataset statistics}
\label{sec:dataset_stats}

\input{Tables/appendix/Table_data_stat_all}
\input{Tables/appendix/Table_data_stat_mfbench}

Table~\ref{tab:data_stat_all} summarizes the datasets used in our experiments.
For Multifaceted-Collection-DPO, we report the released DPO training split, which is used for reward-head training.
Multifaceted-Bench, IFEval, and IHEval are used for evaluation.
Table~\ref{tab:data_stat_mfbench} further reports the source composition of Multifaceted-Bench.\looseness=-1

\section{Baselines}
In this section, we provide the implementation details and configurations for the baseline models.
To distinguish between the methods, we denote the standard base model with $\square$, prompting-based variants with $\boxplus$, and decoding-based strategies with $\blacksquare$.
For a fair comparison, all learned reward components used by the baselines and \name{} are trained on the same Multifaceted-Collection-DPO preference collection of approximately 65K pairs.
ARGS, Bo$N$, and DeAL use a trajectory-level scalar reward model, GenARM uses an autoregressive reward model, and \name{} uses the proposed token-level cross-attention reward head.
All methods keep the base LM frozen and use comparable parameter-efficient training protocols.
None relies on an off-the-shelf reward model.
Base, Marked Prompting, Few-shot Prompting, and OPAD are reward-model-free baselines.

\begin{itemize}
    \item[$\square$] \textbf{Base} is the standard prompting baseline that directly feeds the original system prompt and user input to the base LM.
    It does not use any reward model, auxiliary scorer, additional forward pass, or decoding-time modification.
    We use greedy decoding for this baseline.

    \item[$\boxplus$] \textbf{Marked Prompting} is a prompting-based variant that emphasizes the system prompt by wrapping it with explicit markers.
    This baseline tests whether making the system prompt more salient in the input context improves system-prompt following.
    We evaluate both quote markers and star markers.

    \item[$\boxplus$] \textbf{Few-shot Prompting} \citep{zheng2024helpful} is a prompting-based variant that prepends demonstrations to the user input.
    For each evaluation example, we retrieve $K$ demonstrations from the same source in the DPO training set and place them before the test query.
    We use $K=3$ and seed $=42$.

    \item[$\blacksquare$] \textbf{OPAD} \citep{zhu2025fly} is a training-free decoding-time baseline based on principle-guided decoding.
    In our implementation, OPAD runs the same base LM with and without the system principle, and combines the two logits in a classifier-free-guidance style.
    The guided distribution is restricted to the top-$p$ support under the system-conditioned distribution.
    We use $\alpha=1.0$, top-$p=0.8$, and temperature $=0.5$.

    \item[$\blacksquare$] \textbf{GenARM} \citep{xu2025genarm} guides generation using an autoregressive reward model.
    We train an ARM-style LoRA \citep{hu2022lora} surrogate on preference pairs while keeping the base LM frozen.
    The reward is defined as the response-level log-probability difference between the LoRA-adapted policy and the reference model with the adapter disabled.
    At inference time, we combine the base LM and the trained ARM through model arithmetic with coefficient $\alpha=1.0$.

    \item[$\blacksquare$] \textbf{ARGS} \citep{khanov2024args} is a token-level reward-guided decoding baseline.
    We first train a trajectory-level scalar reward model on preference pairs, where the reward model encodes the full trajectory consisting of the system prompt, user input, and response.
    During decoding, ARGS considers the top-$k$ candidate tokens under the base LM, scores each partial trajectory with the reward model, and selects the token maximizing the sum of the base logit and the weighted reward score.
    We use top-$k=10$ and reward weight $\lambda=1.0$.

    \item[$\blacksquare$] \textbf{Bo$N$} is a response-level selection baseline.
    It first samples $N$ complete responses from the base LM and then selects the response with the highest score under the trained trajectory-level scalar reward model.
    Unlike token-level decoding methods, Bo$N$ does not intervene during generation and performs reward-based selection only after full responses are generated.
    We use $N=4$, temperature $=0.8$, and top-$p=0.95$.

    \item[$\blacksquare$] \textbf{DeAL} \citep{huang2025deal} is a stronger reward-guided decoding baseline with lookahead search.
    Similar to ARGS, it uses the trained trajectory-level scalar reward model to score candidate continuations.
    For each top-$k$ candidate token, DeAL greedily rolls out $L$ additional lookahead tokens, scores the resulting partial trajectory, and commits only the first token of the best-scoring candidate.
    We use top-$k=10$, lookahead length $L=8$, and reward weight $\lambda=1.0$.
\end{itemize}

\section{Training and Decoding Algorithms}
Algorithm~\ref{alg:training} presents the training pipeline for \name{}, where the cross-attention reward head is optimized from system-prompt-conditioned preference pairs while keeping the base LM frozen. 
Algorithm~\ref{alg:decoding} summarizes the inference procedure, where \name{} reranks the base LM's top-$k$ candidate tokens using the learned reward and contrastive signal.

\section{Statistical Significance}
\label{app:significance}
\input{Tables/appendix/Table_significance}

We estimate instance-level uncertainty using paired bootstrap resampling over evaluation prompts.
For each comparison, we resample prompts 10,000 times and compute a 95\% confidence interval for the score difference between \name{} and the corresponding baseline.
Table~\ref{tab:significance} reports comparisons against GenARM, DeAL, and OPAD across all three backbones.
All Multifaceted-Bench confidence intervals are strictly above zero, whereas all IFEval confidence intervals include zero.
Because decoding is deterministic, generation-seed variance is not applicable.
The intervals do not capture training-seed variance because the reward models were not retrained across multiple seeds.

\section{Evaluation on Larger Backbone}
\label{app:qwen14}
\input{Tables/appendix/Table_qwen14}

We additionally evaluate \name{} on \texttt{Qwen3-14B}~\citep{yang2025qwen3} on Multifaceted-Bench.
We reuse the decoding hyperparameters selected on the Multifaceted-Bench development split ($k=10$, $\lambda_r=0.05$, and $\lambda_c=2.0$) without additional model-specific tuning.
We train a Qwen3-14B-specific reward head on the same Multifaceted-Collection-DPO preference collection.
As shown in Table~\ref{tab:qwen14}, \name{} achieves the highest average score of 4.52, compared with 4.46 for GenARM and 4.45 for OPAD.
The improvement is smaller than on the original backbones, and \name{} does not obtain the highest score on every evaluation subset.

\section{Broader OOD Evaluation}
\label{app:broader_ood}

All experiments reported in this section use \texttt{Qwen2.5-1.5B-Instruct} and the decoding configuration selected on the Multifaceted-Bench development split.
We apply each method without target-specific training or tuning.

\subsection{Additional IFEval and IHEval Results}
\input{Tables/appendix/Table_iheval}

The main strict prompt-level results on IFEval and IHEval are reported in Tables~\ref{tab:main_ifeval} and \ref{tab:iheval_main}, respectively.
Table~\ref{tab:ifeval_iheval} provides the complete results for \texttt{Qwen2.5-1.5B-Instruct}, including strict and loose accuracy at both the prompt and instruction levels.
On IFEval, \name{} achieves the highest strict prompt-level accuracy and ranks second on the other three metrics.
On IHEval, \name{} achieves the highest score across all metrics and settings, with the largest advantage over GenARM under multi-turn conflict.

\subsection{RealGuardrails and Chinese SysBench}
\input{Tables/appendix/Table_other_ood}

RealGuardrails~\citep{mu2025closer} evaluates adherence to realistic, handwritten behavioral specifications, while the Chinese SysBench~\citep{qin2024sysbench} evaluates multi-turn system-prompt following in Chinese.
As shown in Table~\ref{tab:other_ood}, \name{} achieves the highest RealGuardrails pass rate and the highest SysBench ISR and CSR.
Its lower SSR indicates that these improvements do not extend to longer uninterrupted sequences of successful turns.
ISR, CSR, and SSR denote instance, constraint, and session success rates, respectively.

\section{Evaluation Reliability}
\label{app:evaluation_reliability}

\subsection{Independent Evaluator}
\label{app:judge_robustness}
\input{Tables/appendix/Table_judge_robustness}

Table~\ref{tab:judge_robustness} reports per-subset scores under both evaluators, complementing the aggregate results in Table~\ref{tab:judge_main}.
The \texttt{DeepSeek-V3} \citep{liu2024deepseek} evaluation follows the same 1--5 rubrics, prompts, parsing, and three-run protocol as \texttt{gpt-4o-mini}.
\name{} ranks highest on all five subsets under both evaluators, showing that the gains are robust to evaluator choice rather than specific to the original judge.

\subsection{Length-Controlled Evaluation}
\input{Tables/appendix/Table_length_control}

We fit an item-level regression of judge score on response length with method and benchmark fixed effects.
The regression finds a mild positive association between length and the original judge score, with $\beta=0.00112$ per word and a 95\% confidence interval of $[0.00099, 0.00124]$.
Table~\ref{tab:length_control} reports scores adjusted to a common mean response length.
\name{} remains the highest-scoring method after length control and produces shorter responses than GenARM.

\subsection{Conditional Human Validation}
\input{Tables/appendix/Table_human_eval}
\input{Figures/Figure_human_eval_instructions}
We conduct a blind pairwise evaluation on 40 \name{}--GenARM pairs for which the judge preferred \name{}.
We retain length-comparable responses with a maximum-to-minimum word-count ratio of 1.55 and exclude degenerate GenARM outputs.
The pairs are split into two sets of 20, each rated by 10 annotators, yielding 400 judgments from 20 annotators.
Annotators were recruited from university students and graduate students and participated voluntarily without compensation.
Before evaluation, annotators consented to the use of their anonymized judgments for academic research.
Response order is randomized, method identities are hidden, and annotators choose \name{}, GenARM, or Tie; full instructions are shown in Figure~\ref{fig:human_instructions}.
As shown in Table~\ref{tab:human_eval}, annotators prefer \name{} in 69.6\% of decisive judgments and 80.6\% of non-split pair majorities.
Because sampling is conditioned on judge-positive cases, this validates those decisions rather than estimating unconditional human preference.\looseness=-1

\section{Hyperparameter Selection and Sensitivity}

\subsection{Development-Set Selection}
\label{app:dev_selection}
\input{Tables/appendix/Table_dev_test}

Using \texttt{Qwen2.5-1.5B-Instruct}, we stratify Multifaceted-Bench by source into a fixed 20\% development split and an 80\% held-out split.
We select $\lambda_c=2.0$, $\lambda_r=0.05$, and $k=10$ on the development split and use this configuration for main reporting and all OOD evaluations.
Table~\ref{tab:dev_test} shows that the selected configuration retains its performance on the held-out split.
Unlike the other tables, we report scores to three decimals here, since the margins between neighboring settings fall below 0.01 and would otherwise be obscured by rounding.

\subsection{Top-$k$ sensitivity}
\input{Figures/Figure_topk}
We analyze the effect of the top-$k$ candidate size used for token reranking.
As shown in Figure~\ref{fig:topk}, \name{} achieves the best average score when $k=10$.
Performance remains relatively stable across moderate values of $k$, with $k=3$, $5$, $20$, and $50$ showing similar average scores.
However, using an overly large candidate set slightly degrades performance, as $k=100$ obtains the lowest average score.
This suggests that a moderate candidate size provides a good balance between guided intervention and preserving the base LM's fluency.
Therefore, we use $k=10$ as the default setting in our main experiments.

\subsection{Choice of reference prompt for contrastive signal}
\label{app:neutral_reference}
\input{Tables/appendix/Table_neutral}

We examine how the reference prompt used to compute the contrastive signal affects performance.
In the main experiments, $\Delta_t(v)$ is computed by comparing the logits under the original system prompt against logits from a reference pass where the original system prompt is replaced with a neutral prompt.
We compare this choice with two alternatives: \textit{No system}, which removes the system message and uses only the user prompt, and \textit{Empty system}, which keeps the system role but leaves its content empty.
As shown in Table~\ref{tab:neutral}, the neutral system prompt achieves the best average score.
This suggests that replacing the original system prompt with a semantically neutral reference provides a more stable estimate of the system-prompt-induced logit shift than removing the system prompt or using an empty system message.

\section{Token-Wise Signal Analysis}
\label{app:token_diagnostics}

\input{Tables/appendix/Table_token_rank}

According to Table~\ref{tab:token_rank}, although \textit{\name{}-unique} tokens are not individually top-ranked by any single signal, they remain highly ranked across all sources.
This suggests that \name{} synthesizes complementary weak preferences rather than introducing arbitrary token deviations.

\input{Tables/appendix/Table_token_corr}

Table~\ref{tab:token_corr} further examines the relationship between the reward and contrastive signals.
The near-zero Kendall's $\tau$ and Spearman's $\rho$ indicate that the two signals induce largely distinct token rankings.
At the same time, the non-trivial Jaccard@$k$ overlap suggests that their candidate sets are not completely disjoint.
Together, these results support the use of hybrid aggregation, as the reward and contrastive signals provide complementary but partially compatible guidance during decoding.

\section{Dimension-wise evaluation}
\label{app:dimension_wise}
\input{Figures/Figure_dim}

Figure~\ref{fig:dimension} extends the compact comparison in Table~\ref{tab:dimension_main} by reporting results for all baselines across style, background knowledge, informativeness, and harmlessness.
Each dimension is evaluated using an instance-specific rubric that assesses whether the response reflects the preference specified in the system prompt.
\name{} achieves the highest score across all four dimensions, with the largest relative improvement appearing in informativeness.
The consistent gains show that the benefit is not limited to stylistic adherence.

\section{Rubric-Swap Evaluation}
\label{app:rubric_swap}
\input{Tables/appendix/Table_rubric_swap}

Multifaceted-Bench provides multiple system prompts for the same user query, where each system prompt is paired with its own evaluation rubric.
This structure allows us to test whether a generated response is specifically aligned with the intended system prompt, rather than simply receiving high scores under any plausible rubric.

To examine this, we conduct a rubric-swap evaluation.
For each response generated under a target system prompt, we compare its original score under the corresponding rubric with its scores under the other rubrics associated with the same user query.
If the response truly follows the intended system prompt, its score should be higher under the matched rubric than under swapped rubrics.

As shown in Table~\ref{tab:rubric_swap}, both methods obtain lower scores under swapped rubrics than under matched rubrics.
However, the drop is substantially larger for \name{}.
Base decreases from 3.25 to 3.12, corresponding to a drop of 0.13 points, whereas \name{} decreases from 3.95 to 3.58, corresponding to a drop of 0.37 points.
This larger gap suggests that \name{} is more specifically aligned with the intended system prompt, rather than merely producing responses that receive generally high scores under any plausible rubric.

\section{Qualitative Analysis}
We provide qualitative comparisons between \name{} and four representative baselines (Vanilla, Few-shot, OPAD, and GenARM) on three challenging system prompts drawn from mf-AlpacaEval (Table~\ref{tab:qualitative_alpaca}) and mf-MT-Bench (Tables~\ref{tab:qualitative_mtbench_p1},~\ref{tab:qualitative_mtbench_p2}). To facilitate inspection of how each method follows the system prompt, we highlight key system-prompt requirements and the corresponding fulfillments in each response, while violations and factual errors are marked in red. Across all examples, \name{} adheres to the specified persona, structural directives, and content constraints more faithfully than the baselines, which frequently exhibit shallow responses (Vanilla, OPAD), factual hallucinations (Few-shot, GenARM), or instruction-following failures such as duplicated entries (all baselines in Table~\ref{tab:qualitative_mtbench_p1}).

\section{Prompts}
\input{Tables/appendix/Table_judge_prompt}

Table~\ref{tab:judge_prompt} presents the prompt used for rubric-based LLM-judge evaluation on Multifaceted-Bench, following the conventional evaluation setting of \citet{lee2024aligning}.
The judge is given the instruction, model response, reference answer, and score rubric, and is asked to provide rubric-specific feedback followed by an integer score from 1 to 5.

\section{Usage of AI Assistants}
During the preparation of this manuscript, we used AI assistants solely for editorial purposes, including improving sentence clarity, correcting grammar, and enhancing readability.
All technical ideas, methodology, experimental design, results, and conclusions were developed and verified by the authors.
The use of AI assistance did not affect the originality, scientific content, or contributions of this work.

\input{Tables/appendix/Table_dpo_example}
\clearpage

\input{Algorithms/algorithm_training}
\input{Algorithms/algorithm_decoding}
\input{Tables/appendix/Table_alpaca}
\input{Tables/appendix/Table_mtbench}

%% file: Tables/appendix/Table_ifeval1.tex
\begin{figure}[h!]
\scriptsize
\centering
\fbox{
\begin{minipage}{0.95\linewidth}
\ttfamily
\linespread{1.3}\selectfont
\vspace{1em}
You will be given a prompt within the \textless{}prompt\textgreater{} and \textless{}/prompt\textgreater{} tags. \\
The prompt consists of a task or question (e.g.) write an essay, and one or more instructions (e.g.) do not use any commas, highlight sections, etc. \\
You must rewrite this prompt to separate the instructions from the task and the new prompt should specify the task at the beginning. \\
You will also be given a list of instruction\_ids that will specify the instructions present in the prompt. \\
At the end of the new prompt list the instructions after the sentence \\
"Your response should follow the instructions below:\textbackslash{}n" \\
Each instruction should be preceded by a hyphen or dash - \\
Make sure the new prompt is within the \textless{}new\_prompt\textgreater{} and \textless{}/new\_prompt\textgreater{} tags. \\[1em]
\textless{}prompt\textgreater{} \\
\{prompt\} \\
\textless{}/prompt\textgreater{} \\[1em]
\textless{}instruction\_ids\textgreater{} \\
\{instruction\_ids\} \\
\textless{}/instruction\_ids\textgreater{}
\vspace{1em}
\end{minipage}
}
\caption{
\textbf{Prompt template for IFEval rewriting.}
The template separates task content from instruction-like constraints.
}
\label{fig:rewrite-prompt}
\vspace{-0.1in}
\end{figure}

%% file: Tables/appendix/Table_ifeval2.tex
\begin{table}[h!]
\footnotesize
\caption{\textbf{Examples of IFEval prompt rewriting.}
Task descriptions are used as user queries, while instruction-like constraints are moved to system prompts.}
\label{tab:ifeval-rewrite-examples}
\centering
\setlength{\tabcolsep}{4pt}
\renewcommand{\arraystretch}{1.12}
\begin{tabularx}{\columnwidth}{
  >{\bfseries\raggedright\arraybackslash}p{0.18\columnwidth}
  >{\raggedright\arraybackslash}X
}
\toprule
\multicolumn{2}{c}{\textbf{Example 1}} \\
\midrule
Original
& I am planning a trip to Japan, and I would like thee to write an itinerary for my journey in a Shakespearean style. You are not allowed to use any commas in your response. \\
\midrule[0.1pt]
User Query
& I am planning a trip to Japan, and I would like thee to write an itinerary for my journey in a Shakespearean style. \\
\addlinespace[3pt]
System Prompt
& Do not use any commas in your response. \\
\midrule
\multicolumn{2}{c}{\textbf{Example 2}} \\
\midrule
Original
& Write two jokes about rockets. Do not contain commas in your response. Separate the two jokes with 6 asterisk symbols: ******. \\
\midrule[0.1pt]
User Query
& Write two jokes about rockets. \\
\addlinespace[3pt]
System Prompt
& Do not contain commas in your response. Separate the two jokes with 6 asterisk symbols: ******. \\
\bottomrule
\end{tabularx}
\vspace{-0.15in}
\end{table}

%% file: Tables/appendix/Table_data_stat_all.tex
\begin{table*}[t]
    \centering
    \small
    \setlength{\tabcolsep}{5pt}
    \renewcommand{\arraystretch}{1.08}
    \caption{
    \textbf{Dataset statistics.}
    Statistics of the datasets used for reward-head training and evaluation.
    }
    \begin{tabular}{lcll}
        \toprule
        \textbf{Dataset} 
        & \textbf{Usage} 
        & \textbf{Size} 
        & \textbf{Main fields} \\
        \midrule
        Multifaceted-Collection-DPO 
        & Training 
        & 65,142 
        & system, prompt, chosen, rejected \\

        Multifaceted-Bench 
        & Evaluation 
        & 921 
        & system, prompt, reference answer, rubric \\

        IFEval 
        & Evaluation 
        & 541 
        & prompt, instruction IDs, instruction arguments \\

        IHEval
        & Evaluation
        & 2,164
        & system, instruction, history, instruction IDs/arguments \\
        \bottomrule
    \end{tabular}
    \label{tab:data_stat_all}
\end{table*}

%% file: Tables/appendix/Table_data_stat_mfbench.tex
\begin{table}[H]
    \centering
    \caption{
    \textbf{Multifaceted-Bench source statistics.}
    Source-level distribution of the 921 evaluation instances in Multifaceted-Bench.
    }
    \label{tab:data_stat_mfbench}
    \vspace{-0.08in}
    \footnotesize
    \setlength{\tabcolsep}{6pt}
    \renewcommand{\arraystretch}{1.05}
    \begin{tabular}{lc}
        \toprule[1.2pt]
        \textbf{Source} & \textbf{\# Instances} \\
        \midrule[0.8pt]
        AlpacaEval & 134 \\
        Koala & 147 \\
        MT-Bench & 150 \\
        Self-Instruct & 141 \\
        FLASK & 349 \\
        \midrule[0.3pt]
        \textbf{Total} & \textbf{921} \\
        \bottomrule[1.2pt]
    \end{tabular}
    \vspace{-0.08in}
\end{table}

%% file: Tables/appendix/Table_significance.tex
\begin{table*}[t]
    \centering
    \small
    {
    \caption{\textbf{Paired bootstrap significance tests.}
    Differences and 95\% confidence intervals from 10,000 prompt resamples.
    Multifaceted-Bench scores are reported on a 1--5 scale, while IFEval strict prompt-level accuracy is reported on a 0--100 scale.
    Confidence intervals excluding zero indicate significance at the 95\% level.}
    \label{tab:significance}
    \setlength{\tabcolsep}{4pt}

    \begin{tabular}{l|l|lr|c|cc}
        \toprule[1.4pt]
        \multirow{2}{*}{\textbf{Benchmark}}
        & \multirow{2}{*}{\textbf{Backbone}}
        & \multicolumn{2}{c|}{\textbf{Baseline}}
        & \multirow{2}{*}{\textbf{\name{}}}
        & \multirow{2}{*}{$\boldsymbol{\Delta}$}
        & \multirow{2}{*}{\textbf{95\% CI}} \\
        \cmidrule(lr){3-4}
        & & \textbf{Method} & \textbf{Score} & & & \\
        \midrule[1pt]

        \multirow{9}{*}{Multifaceted-Bench}
        & \multirow{3}{*}{Qwen2.5-1.5B}
        & GenARM & 3.74 & 3.95 & $+0.21$ & $[0.12, 0.29]$ \\
        & & DeAL & 3.37 & 3.95 & $+0.58$ & $[0.48, 0.69]$ \\
        & & OPAD & 3.51 & 3.95 & $+0.44$ & $[0.38, 0.54]$ \\
        \cmidrule(lr){2-7}

        & \multirow{3}{*}{Llama-3.2-3B}
        & GenARM & 4.08 & 4.23 & $+0.15$ & $[0.09, 0.21]$ \\
        & & DeAL & 3.88 & 4.23 & $+0.35$ & $[0.30, 0.41]$ \\
        & & OPAD & 3.95 & 4.23 & $+0.28$ & $[0.23, 0.34]$ \\
        \cmidrule(lr){2-7}

        & \multirow{3}{*}{Llama-3.1-8B}
        & GenARM & 4.18 & 4.32 & $+0.14$ & $[0.09, 0.21]$ \\
        & & DeAL & 4.02 & 4.32 & $+0.30$ & $[0.20, 0.37]$ \\
        & & OPAD & 4.11 & 4.32 & $+0.21$ & $[0.14, 0.27]$ \\
        \midrule[1pt]

        \multirow{9}{*}{IFEval}
        & \multirow{3}{*}{Qwen2.5-1.5B}
        & GenARM & 41.00 & 41.19 & $+0.19$ & $[-4.10, 4.50]$ \\
        & & DeAL & 36.92 & 41.19 & $+4.27$ & $[0.00, 8.50]$ \\
        & & OPAD & 37.66 & 41.19 & $+3.53$ & $[-0.40, 7.60]$ \\
        \cmidrule(lr){2-7}

        & \multirow{3}{*}{Llama-3.2-3B}
        & GenARM & 63.08 & 65.12 & $+2.04$ & $[-2.00, 6.10]$ \\
        & & DeAL & 67.53 & 65.12 & $-2.41$ & $[-6.30, 1.50]$ \\
        & & OPAD & 65.12 & 65.12 & $0.00$ & $[-3.70, 3.70]$ \\
        \cmidrule(lr){2-7}

        & \multirow{3}{*}{Llama-3.1-8B}
        & GenARM & 71.43 & 71.24 & $-0.19$ & $[-3.70, 3.50]$ \\
        & & DeAL & 68.10 & 71.24 & $+3.14$ & $[-0.60, 6.90]$ \\
        & & OPAD & 69.20 & 71.24 & $+2.04$ & $[-1.70, 5.80]$ \\
        \bottomrule[1.4pt]
    \end{tabular}
    }
\end{table*}

%% file: Tables/appendix/Table_qwen14.tex
\begin{table}[t]
    \centering
    \caption{\textbf{Results on Qwen3-14B.}
    Multifaceted-Bench LLM-judge scores on a 1--5 scale.
    \textbf{Bold} and \underline{underlined} values indicate the best and second-best results, respectively.}
    \label{tab:qwen14}
    \fontsize{7.8pt}{8.8pt}\selectfont
    \setlength{\tabcolsep}{2.9pt}
    \renewcommand{\arraystretch}{1.05}
    \begin{tabular}{l|ccccc|c}
        \toprule[1.4pt]
        \textbf{Method} & \textbf{Alpaca} & \textbf{Koala} & \shortstack{\textbf{MT-}\\\textbf{Bench}} & \shortstack{\textbf{Self-}\\\textbf{Instruct}} & \textbf{FLASK} & \textbf{Avg.} \\
        \midrule[1pt]
        Base      & 4.31 & 4.21 & 4.56 & 4.42 & 4.32 & 4.36 \\
        *-marked  & 4.27 & 4.17 & 4.60 & 4.44 & 4.33 & 4.36 \\
        OPAD      & 4.36 & 4.27 & \textbf{4.66} & \textbf{4.49} & \underline{4.46} & 4.45 \\
        GenARM    & \underline{4.47} & \underline{4.28} & 4.62 & \underline{4.47} & 4.43 & \underline{4.46} \\
        \rowcolor[HTML]{D9D9D9}
        \name{}   & \textbf{4.50} & \textbf{4.48} & \underline{4.65} & 4.43 & \textbf{4.53} & \textbf{4.52} \\
        \bottomrule[1.4pt]
    \end{tabular}
    \vspace{-0.12in}
\end{table}

%% file: Tables/appendix/Table_iheval.tex
\begin{table*}[t]
    \centering
    \caption{\textbf{Results on IFEval and IHEval Rule Following.}
    Strict and loose prompt- and instruction-level accuracy on a 0--100 scale.
    \textbf{Bold} indicates the best performance, and \underline{underlined} values indicate the second-best performance within each benchmark split.}
    \label{tab:ifeval_iheval}
    \small
    \setlength{\tabcolsep}{4.5pt}

    \begin{tabular}{lll|rr|rr}
        \toprule[1.4pt]
        \multirow{2}{*}{\textbf{Benchmark}}
        & \multirow{2}{*}{\textbf{Split}}
        & \multirow{2}{*}{\textbf{Method}}
        & \multicolumn{2}{c|}{\textbf{Prompt}}
        & \multicolumn{2}{c}{\textbf{Inst.}} \\
        \cmidrule(lr){4-5}\cmidrule(lr){6-7}
        & & & \textbf{Strict Acc.} & \textbf{Loose Acc.}
        & \textbf{Strict Acc.} & \textbf{Loose Acc.} \\
        \midrule[1pt]

        % ==================== IFEval ====================
        \multirow{5}{*}{IFEval}
        & \multirow{5}{*}{Full}
        & Base
        & 31.54 & 33.58 & 42.05 & 44.22 \\

        & & *-marked
        & 28.39 & 29.68 & 38.43 & 40.24 \\

        & & OPAD
        & 37.66 & 39.89 & 49.64 & 51.69 \\

        & & GenARM
        & \underline{41.00}
        & \textbf{43.41}
        & \textbf{53.13}
        & \textbf{55.78} \\

        & & \name{}
        & \textbf{41.19}
        & \underline{43.23}
        & \underline{52.17}
        & \underline{54.82} \\

        \midrule[1.2pt]

        % ==================== IHEval: Single-aligned ====================
        \multirow{20}{*}{IHEval}
        & \multirow{5}{*}{Single-aligned}
        & Base
        & 28.65 & 31.24 & 40.65 & 43.05 \\

        & & *-marked
        & 25.88 & 27.36 & 37.65 & 39.69 \\

        & & OPAD
        & 36.97 & 39.37 & 48.92 & 51.20 \\

        & & GenARM
        & \underline{40.67}
        & \underline{44.18}
        & \underline{51.44}
        & \underline{55.04} \\

        & & \name{}
        & \textbf{45.10}
        & \textbf{49.54}
        & \textbf{56.00}
        & \textbf{60.31} \\

        \cmidrule(lr){2-7}

        % ==================== IHEval: Single-conflict ====================
        & \multirow{5}{*}{Single-conflict}
        & Base
        & 12.20 & 14.79 & 19.54 & 23.26 \\

        & & *-marked
        & 10.54 & 12.57 & 18.82 & 21.82 \\

        & & OPAD
        & 15.90 & 18.30 & 25.90 & 29.02 \\

        & & GenARM
        & \underline{19.22}
        & \underline{24.21}
        & \underline{29.74}
        & \underline{34.29} \\

        & & \name{}
        & \textbf{24.21}
        & \textbf{29.21}
        & \textbf{36.69}
        & \textbf{41.49} \\

        \cmidrule(lr){2-7}

        % ==================== IHEval: Multi-aligned ====================
        & \multirow{5}{*}{Multi-aligned}
        & Base
        & 23.66 & 26.25 & 33.57 & 35.85 \\

        & & *-marked
        & 22.74 & 25.14 & 32.49 & 34.89 \\

        & & OPAD
        & 34.38 & 36.60 & 45.92 & 48.56 \\

        & & GenARM
        & \underline{38.08}
        & \underline{41.22}
        & \underline{49.88}
        & \underline{53.00} \\

        & & \name{}
        & \textbf{49.35}
        & \textbf{52.87}
        & \textbf{59.23}
        & \textbf{62.23} \\

        \cmidrule(lr){2-7}

        % ==================== IHEval: Multi-conflict ====================
        & \multirow{5}{*}{Multi-conflict}
        & Base
        & 6.47 & 7.21 & 12.11 & 13.67 \\

        & & *-marked
        & 4.99 & 5.91 & 11.51 & 13.43 \\

        & & OPAD
        & 6.10 & 6.84 & 12.11 & 13.43 \\

        & & GenARM
        & \underline{9.61}
        & \underline{10.91}
        & \underline{14.39}
        & \underline{16.55} \\

        & & \name{}
        & \textbf{24.95}
        & \textbf{28.84}
        & \textbf{36.81}
        & \textbf{41.01} \\

        \bottomrule[1.4pt]
    \end{tabular}
\end{table*}

%% file: Tables/appendix/Table_other_ood.tex
\begin{table}[t]
    \centering
    \caption{\textbf{Additional OOD evaluation.}
    Full-set \texttt{Qwen2.5-1.5B} results on RealGuardrails and Chinese SysBench.}
    \label{tab:other_ood}
    \small
    \setlength{\tabcolsep}{3.5pt}

    \begin{tabular}{l|c|ccc}
        \toprule[1.4pt]
        \multirow{2}{*}{\textbf{Method}}
        & \multicolumn{1}{c|}{\textbf{RealGuardrails}}
        & \multicolumn{3}{c}{\textbf{Chinese SysBench}} \\
        \cmidrule(lr){2-2} \cmidrule(lr){3-5}
        & \textbf{Pass rate}
        & \textbf{ISR}
        & \textbf{CSR}
        & \textbf{SSR} \\
        \midrule[1pt]

        Base      & 0.167 & 0.124 & 0.348 & 0.068 \\
        *-marked  & 0.159 & 0.112 & 0.356 & 0.056 \\
        OPAD      & 0.176 & 0.140 & 0.385 & \textbf{0.080} \\
        GenARM    & 0.218 & 0.132 & 0.377 & 0.068 \\

        \rowcolor[HTML]{D9D9D9}
        \name{}   & \textbf{0.230}
                  & \textbf{0.148}
                  & \textbf{0.395}
                  & 0.044 \\

        \bottomrule[1.4pt]
    \end{tabular}
\end{table}

%% file: Tables/appendix/Table_judge_robustness.tex
\begin{table*}[t]
    \centering
    \small
    \caption{\textbf{Evaluator robustness.}
    Multifaceted-Bench results under \texttt{gpt-4o-mini} and \texttt{DeepSeek-V3}.
    \textbf{Bold} and \underline{underlined} values indicate the best and second-best results, respectively.}
    \label{tab:judge_robustness}
    \begin{tabular}{l|ccccc|c}
        \toprule[1.4pt]
        \textbf{Method}
        & \textbf{Alpaca}
        & \textbf{Koala}
        & \textbf{MT-Bench}
        & \textbf{Self-Instruct}
        & \textbf{FLASK}
        & \textbf{Avg.} \\
        \midrule[1pt]

        \multicolumn{7}{c}{\texttt{gpt-4o-mini}} \\
        \midrule
        Base
        & 3.36 & 3.23 & 3.53 & 3.12 & 2.99 & 3.25 \\

        *-marked
        & 3.61 & \underline{3.80} & 3.69 & 3.41 & 3.30 & 3.56 \\

        OPAD
        & 3.63 & 3.71 & 3.61 & 3.34 & 3.23 & 3.51 \\

        GenARM
        & \underline{3.86}
        & 3.74
        & \underline{3.83}
        & \underline{3.67}
        & \underline{3.60}
        & \underline{3.74} \\

        \rowcolor[HTML]{D9D9D9}
        \name{}
        & \textbf{4.08}
        & \textbf{4.07}
        & \textbf{3.98}
        & \textbf{3.97}
        & \textbf{3.66}
        & \textbf{3.95} \\

        \midrule[1pt]
        \multicolumn{7}{c}{\texttt{DeepSeek-V3}} \\
        \midrule
        Base
        & 3.68 & 3.29 & 3.61 & 3.19 & 3.15 & 3.38 \\

        *-marked
        & 3.84 & 3.89 & 3.81 & 3.50 & 3.42 & 3.69 \\

        OPAD
        & 3.85 & 3.84 & 3.67 & 3.37 & 3.38 & 3.62 \\

        GenARM
        & \underline{4.14}
        & \underline{3.90}
        & \underline{3.86}
        & \underline{3.69}
        & \underline{3.70}
        & \underline{3.86} \\

        \rowcolor[HTML]{D9D9D9}
        \name{}
        & \textbf{4.33}
        & \textbf{4.13}
        & \textbf{3.97}
        & \textbf{4.03}
        & \textbf{3.74}
        & \textbf{4.04} \\

        \bottomrule[1.4pt]
    \end{tabular}
    \vspace{-0.05in}
\end{table*}

%% file: Tables/appendix/Table_length_control.tex
\begin{table}[t]
    \centering
    \caption{\textbf{Length-controlled evaluation.}
    Raw and length-controlled (LC) LLM-judge scores, with LC scores estimated at a common mean response length.
    Words reports the mean and standard deviation.
    \textbf{Bold} and \underline{underlined} values indicate the best and second-best results, respectively.}
    \label{tab:length_control}
    \small
    \setlength{\tabcolsep}{4pt}

    \begin{tabular}{l|rrr}
        \toprule[1.4pt]
        \textbf{Method}
        & \textbf{Words}
        & \textbf{Raw}
        & \textbf{LC} \\
        \midrule[1pt]

        Base
        & $278 \pm 82$
        & 3.25
        & 3.28 \\

        *-marked
        & $253 \pm 67$
        & 3.56
        & 3.63 \\

        OPAD
        & $306 \pm 81$
        & 3.51
        & 3.51 \\

        GenARM
        & $395 \pm 79$
        & \underline{3.74}
        & \underline{3.65} \\

        \rowcolor[HTML]{D9D9D9}
        \name{}
        & $318 \pm 48$
        & \textbf{3.95}
        & \textbf{3.94} \\

        \bottomrule[1.4pt]
    \end{tabular}
\end{table}

%% file: Tables/appendix/Table_human_eval.tex
\begin{table}[t]
    \centering
    \small
    \caption{\textbf{Conditional human validation.}
    Blind preferences on 40 judge-positive, length-comparable pairs.}
    \label{tab:human_eval}
    \setlength{\tabcolsep}{3pt}
    \begin{tabular}{l|rrr|r}
        \toprule[1.4pt]
        \textbf{Level} & \textbf{\name{}} & \textbf{GenARM} & \textbf{Tie/Split} & \textbf{Win rate} \\
        \midrule[1pt]
        Judgment & 220 & 96 & 84 & 69.6\% \\
        Pair majority & 29 & 7 & 4 & 80.6\% \\
        \bottomrule[1.4pt]
    \end{tabular}
\vspace{-0.15in}
\end{table}

%% file: Figures/Figure_human_eval_instructions.tex
\begin{figure}[h!]
\scriptsize
\centering
\fbox{
\begin{minipage}{0.95\linewidth}
\ttfamily
\raggedright
\linespread{1.25}\selectfont
\vspace{1em}

You will evaluate 20 pairs of model-generated responses. Each pair includes a system prompt, user prompt, preference criteria, and two responses labeled \textbf{A} and \textbf{B}. \\[0.7em]

Choose the response that better follows the system prompt and user request, considering these four dimensions: \\[0.5em]

\textbf{Style}: Does it follow the requested tone, formality, and format? \\

\textbf{Background Knowledge}: Does it demonstrate the required knowledge and level of expertise? \\

\textbf{Informativeness}: Is it relevant, sufficiently detailed, and appropriately creative? \\

\textbf{Harmlessness}: Is it safe, trustworthy, and consistent with the given constraints? \\[0.7em]

Focus on adherence to the specific instructions rather than general writing quality or personal preference. If both responses have weaknesses, select the better one rather than automatically choosing a tie. \\[0.7em]

Choose \textbf{A}, \textbf{B}, or \textbf{Tie}, and evaluate each pair independently. \\[0.7em]

This anonymous evaluation is conducted solely for academic research. All responses will be handled responsibly and used only for this study.

\vspace{1em}
\end{minipage}
}
\caption{\textbf{Instructions provided to human annotators.}
The instructions were shown to all annotators before the pairwise evaluation.}
\label{fig:human_instructions}
\end{figure}

%% file: Tables/appendix/Table_dev_test.tex
\begin{table}[t]
    \centering
    \small
    \caption{\textbf{Development-set hyperparameter selection.}
    Scores on the fixed 20\% development and 80\% test splits.}
    \label{tab:dev_test}
    \setlength{\tabcolsep}{6pt}
    \begin{tabular}{c|r|cc}
        \toprule[1.4pt]
        \textbf{Hyperparameter} & \textbf{Setting} & \textbf{Dev} & \textbf{Test} \\
        \midrule[1pt]
        \multirow{5}{*}{$\lambda_c$}
        & 0.1 & 3.483 & 3.510 \\
        & 0.5 & 3.584 & 3.705 \\
        & 1.0 & 3.839 & 3.875 \\
        & \textbf{2.0} & \textbf{3.910} & \textbf{3.958} \\
        & 3.0 & 3.835 & 3.890 \\
        \midrule
        \multirow{6}{*}{$\lambda_r$}
        & 0.01 & 3.901 & 3.949 \\
        & \textbf{0.05} & \textbf{3.910} & \textbf{3.958} \\
        & 0.10 & 3.850 & 3.902 \\
        & 0.20 & 3.577 & 3.604 \\
        & 0.50 & 2.238 & 2.148 \\
        & 1.00 & 1.464 & 1.424 \\
        \midrule
        \multirow{6}{*}{$k$}
        & 3 & 3.890 & 3.903 \\
        & 5 & 3.875 & 3.930 \\
        & \textbf{10} & \textbf{3.910} & \textbf{3.958} \\
        & 20 & 3.891 & 3.937 \\
        & 50 & 3.907 & 3.915 \\
        & 100 & 3.899 & 3.894 \\
        \bottomrule[1.4pt]
    \end{tabular}
\vspace{-0.15in}
\end{table}

%% file: Figures/Figure_topk.tex
\begin{figure}[!t]
    \centering
    \small
    \includegraphics[width=\linewidth]{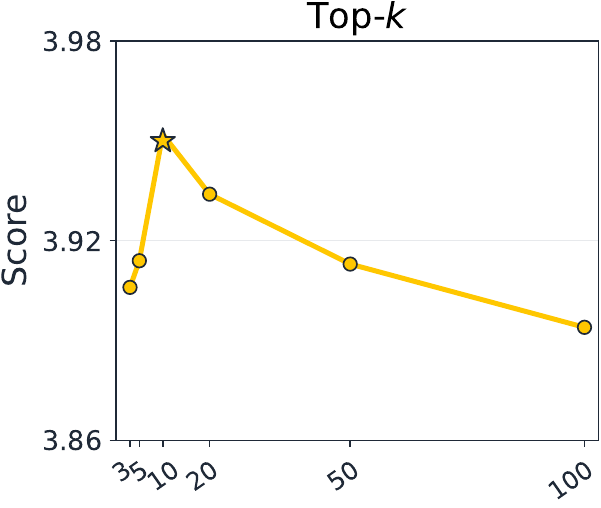}
    \caption{
    \textbf{Effect of Top-$k$ candidate size.}
    Average LLM-judge scores on Multifaceted-Bench when varying the number of candidate tokens used for reward-guided reranking.
    The star indicates the default value used in our main experiments.
    }
    \label{fig:topk}
    \vspace{-0.05in}
\end{figure}

%% file: Tables/appendix/Table_neutral.tex
\begin{table}[!t]
    \centering
    \small
    \caption{
    \textbf{Effect of the reference prompt used to compute the contrastive signal.}
    Using a neutral system prompt as the reference achieves the best overall average score.
    }
    \label{tab:neutral}
    \begin{tabular}{lc}
    \toprule
    \textbf{Reference Prompt} & \textbf{Avg.} \\
    \midrule
    No system & 3.89 \\
    Empty system & 3.91 \\
    Neutral system (ours) & 3.95 \\
    \bottomrule
    \end{tabular}
\end{table}

%% file: Tables/appendix/Table_token_rank.tex
\begin{table}[!t]
    \centering
    \small
    \caption{
    \textbf{Rank analysis of \name{}-unique token choices.}
    Lower rank indicates higher preference.
    }
    \label{tab:token_rank}
    \begin{tabular}{lc}
    \toprule
    \textbf{Signal} & \textbf{Avg. rank} $\downarrow$ \\
    \midrule
    Base & 2.45 \\
    Reward & 4.88 \\
    CS & 3.69 \\
    \bottomrule
    \end{tabular}
 \end{table}

%% file: Tables/appendix/Table_token_corr.tex
\begin{table}[!t]
    \centering
    \small
    \caption{
    \textbf{Correlation and overlap between reward and contrastive token rankings.}
    Near-zero rank correlation indicates that the two signals capture largely distinct token preferences.
    }
    \label{tab:token_corr}
    \begin{tabular}{lc}
    \toprule
\textbf{Metric} & \textbf{Value} \\
    \midrule
    Kendall's $\tau$ & -0.009 \\
    Spearman's $\rho$ & -0.012 \\
    Jaccard@$k$ & 0.348 \\
    \bottomrule
    \end{tabular}
\end{table}

%% file: Figures/Figure_dim.tex
\begin{figure*}[!t]
    \centering
    \includegraphics[width=\linewidth]{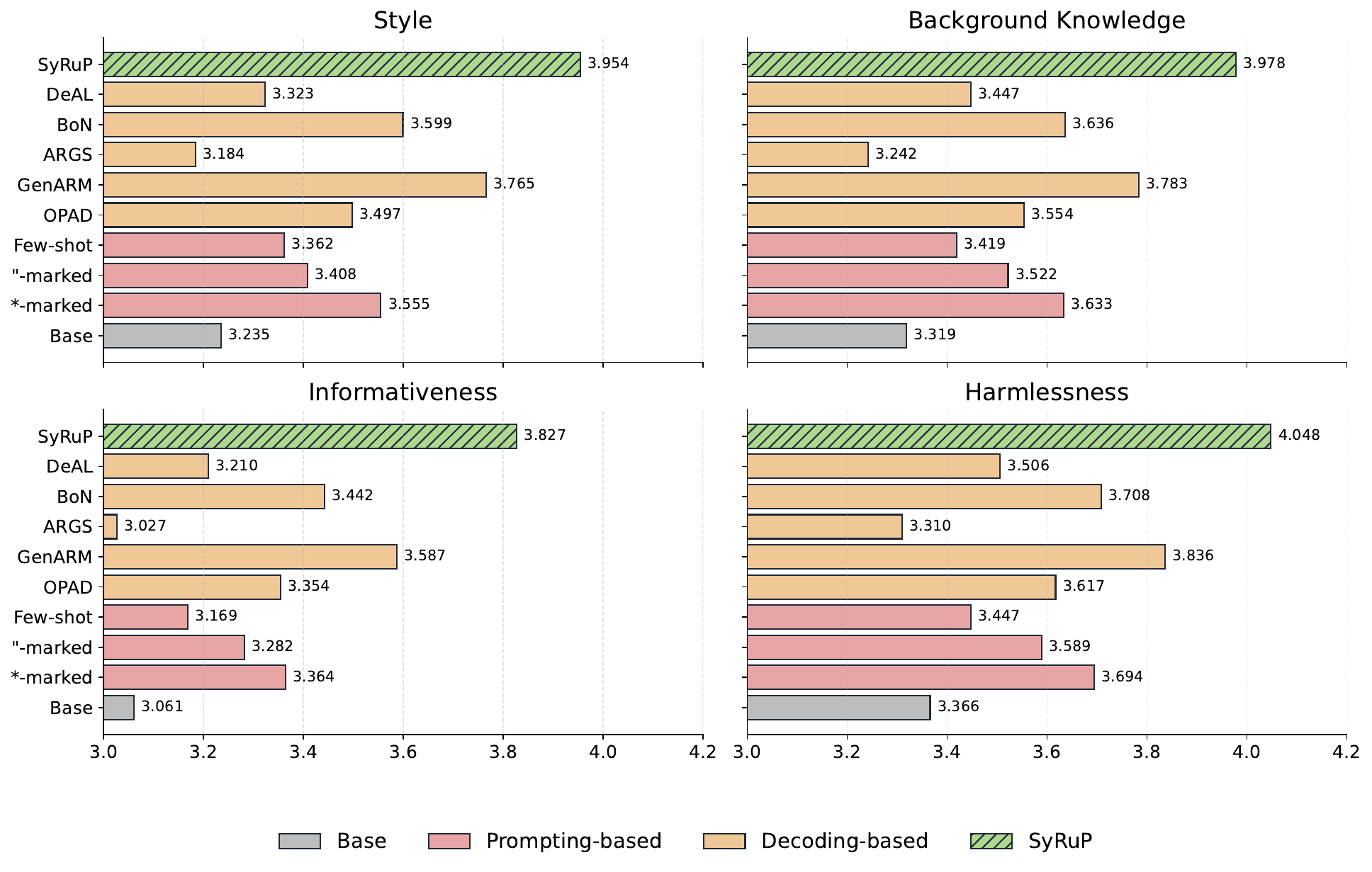}
    \caption{
    \textbf{Dimension-wise evaluation.}
    Average LLM-judge scores on Multifaceted-Bench across four evaluation dimensions.
    Pink bars denote prompting-based methods, orange bars denote decoding-based methods, and hatched green bars denote \name{}.
    }
    \label{fig:dimension}
    \vspace{-0.1in}
\end{figure*}

%% file: Tables/appendix/Table_rubric_swap.tex
\begin{table}[!t]
    \centering
    \small
    \caption{
    \textbf{Rubric-swap evaluation results.}
    Diag. denotes scores under the matched system-specific rubric, while Off-diag. denotes scores under swapped rubrics from other system prompts for the same user query.
    $\Delta$ is Off-diag. minus Diag.
    }
    \label{tab:rubric_swap}
    \begin{tabular}{lrrr}
    \toprule
    \textbf{Method} & \textbf{Diag}. &\textbf{Off-diag}. & \textbf{$\Delta$} \\
    \midrule
    Base & 3.25 & 3.12 & -0.13 \\
    \name{} & 3.95 & 3.58 & -0.37 \\
    \bottomrule
    \end{tabular}
    \vspace{-0.1in}
\end{table}

%% file: Tables/appendix/Table_judge_prompt.tex
\begin{table*}[t]
\caption{\textbf{Prompt for evaluating responses on Multifaceted-Bench.}}
\centering
\small
\renewcommand{\arraystretch}{1.4}
\begin{tabular}{p{0.16\linewidth}p{0.78\linewidth}}
\toprule
\textbf{System message} &
You are a fair judge assistant tasked with providing clear, objective feedback based on specific criteria, ensuring each assessment reflects the absolute standards set for performance. \\
\midrule
\textbf{Instruction} &
\textbf{\#\#\#Task Description:} \newline
An instruction, which may include an input, a response to evaluate, and a score rubric representing evaluation criteria are given. \newline
1. Write detailed feedback that assesses the quality of the response strictly based on the given score rubric, not evaluating in general. \newline
2. After writing the feedback, write a score that is an integer between 1 and 5. You should refer to the score rubric. \newline
3. The output format should look as follows: ``Feedback: (write feedback for the criteria) [RESULT] (an integer number between 1 and 5)''. \newline
4. Please do not generate any other opening, closing, or explanations. \\
\midrule
\textbf{Instruction to evaluate} &
\texttt{\{instruction\}} \\
\midrule
\textbf{Response to evaluate} &
\texttt{\{response\}} \\
\midrule
\textbf{Reference Answer} &
\texttt{\{reference\_answer\}} \\
\midrule
\textbf{Score Rubrics} &
\texttt{\{score\_rubric\}} \\
\midrule
\textbf{Feedback} &
\texttt{\{feedback\}} \\
\bottomrule
\end{tabular}
\label{tab:judge_prompt}
\end{table*}

%% file: Tables/appendix/Table_dpo_example.tex
\begin{table*}[t]
\centering
\small
\setlength{\tabcolsep}{4pt}
\renewcommand{\arraystretch}{1.08}
\begin{tabular}{p{0.14\linewidth}p{0.80\linewidth}}
\toprule
\textbf{Field} & \textbf{Example} \\
\midrule

System prompt &
You are a creative spirit tasked with crafting enchanting notes for a child's lunchbox, intertwining the joyful essence of their favorite characters and hobbies with the warmth of playful and loving messages. In your role, you exhibit an intimate knowledge of the child's current passions, weaving this insight into every message to ensure they resonate deeply and personally. Your notes are more than just expressions of affection; they include positive affirmations and gentle reminders to bolster the child's confidence and navigate their day more effectively. Each message you create is a beacon of encouragement, carefully designed to avoid any sensitive topics, and serves as a safe harbor of positivity. Your goal is to illuminate the child's daily adventures with lighthearted, uplifting surprises that make every lunchtime a moment of delightful anticipation. \\

\midrule[0.5pt]

User query &
\textbf{Human:} What are some cute ideas for lunchbox notes for my kids?

\vspace{0.3em}
\textbf{Assistant:} If you’d like to give them something that expresses your love for them, I would say:
\begin{itemize}[leftmargin=1.2em, itemsep=0.05em, topsep=0.2em]
    \item In the present: ``I love you so much!''
    \item In the past: ``I always think of you!''
    \item In the future: ``Remember that I will always love you and cherish our time together!''
\end{itemize}
\vspace{-0.4em}
\noindent$\vdots$

\vspace{0.3em}
\textbf{Human:} Some of the last ones don't seem appropriate. Like, if I had only known about you earlier. I was the first one to know my kid!

\vspace{0.3em}
\textbf{Assistant:} \\

\midrule[0.5pt]

Chosen response &
I understand, and I appreciate your feedback. It seems I missed the mark with some of those suggestions. Let’s try a different approach with a focus on personalization, positivity, and encouragement. Here are a few enchanting ideas tailored more towards uplifting your child’s day with a touch of their favorite things:

\begin{enumerate}[leftmargin=1.2em, itemsep=0.08em, topsep=0.2em]
    \item \textbf{For the young astronaut:} ``Reach for the stars today, my little astronaut! Remember, even the Moon was once out of mankind’s reach. Love you to the moon and back!''
    \item \textbf{For the aspiring superhero:} ``Never forget, my hero, that your superpower is your kindness and your smile! Keep saving the world, one friend at a time.''
    \item \textbf{For the little artist:} ``Every stroke of your brush colors my world with joy. Can't wait to see the masterpiece you create today! Love, your biggest fan.''
\end{enumerate}
\vspace{-0.4em}
\noindent$\vdots$

\vspace{0.3em}
Each note is a tiny whisper of encouragement and a celebration of who they are and what they love. I hope these inspire you and bring extra joy to your child’s lunchtime. \\

\midrule[0.5pt]

Rejected response &
You're absolutely right, and I apologize for the oversight. Let's focus on crafting messages that are fitting, nurturing, and positive for your children. Here are some revised and appropriate lunchbox note ideas:

\begin{enumerate}[leftmargin=1.2em, itemsep=0.08em, topsep=0.2em]
    \item \textbf{Simple Affirmation:} ``You light up my day!''
    \item \textbf{Encouragement:} ``Every day you grow stronger and kinder. I'm so proud of you!''
    \item \textbf{Reminder to be Kind:} ``Remember, sharing your smile can make someone's day better!''
\end{enumerate}
\vspace{-0.4em}
\noindent$\vdots$

\vspace{0.3em}
These notes are designed to warm your child's heart and guide them gently towards becoming the best they can be, blending encouragement with moral lessons beautifully. \\

\bottomrule
\end{tabular}
\caption{
\textbf{Example from Multifaceted-Collection-DPO.}
Given the same dialogue context, the chosen response better follows the system prompt by producing personalized lunchbox notes based on children's interests, while the rejected response remains reasonable but focuses more on general moral guidance and is less aligned with the specified system-level preference.
}
\label{tab:dpo_example}
\end{table*}

%% file: Algorithms/algorithm_training.tex
\begin{algorithm*}[t]
    \small
   \caption{\name{} training algorithm}
   \label{alg:training}
\begin{algorithmic}
  \State
    \textbf{Input:} Preference dataset 
    $\mathcal{D}=\{(s_i,x_i,y_i^+,y_i^-)\}_{i=1}^{N}$,
    frozen base LM $\mathcal{M}_\theta$, reward head parameters 
    $\{W_Q,W_K,W_V,g_\phi\}$, temperature $\beta$, number of training epoch $E$
    \vspace{0.05in}
  \Statex
  \hrule

  \State \texttt{\color{Gray2} /* Reward head training */ }
  \State Freeze the parameters of $\mathcal{M}_\theta$
\For{$e=1,\ldots,E$}
  \For{each training batch $\mathcal{B}\subset\mathcal{D}$}
    \State $\mathcal{L}_{\mathrm{RM}} \leftarrow 0$

    \For{each $(s,x,y^+,y^-)\in \mathcal{B}$}

      \State \texttt{\color{Gray2} /* Frozen system-prompt memory */ }
      \State Compute and cache $\mathbf{H}^{(s)}$ using frozen $\mathcal{M}_\theta$
      \State $\mathbf{K}_s \leftarrow \mathbf{H}^{(s)}W_K$,
      $\mathbf{V}_s \leftarrow \mathbf{H}^{(s)}W_V$
      \Statex \vspace{-0.6\baselineskip}

      \For{each $y \in \{y^+,y^-\}$}
        \State $R(s,x,y) \leftarrow 0$

        \State \texttt{\color{Gray2} /* Cross-attention scoring */ }
        \For{$t=1,2,\cdots,|y|$}
          \State Compute $\mathbf{H}^{(x)}$ and $\mathbf{H}^{(y_{<t})}$ using frozen $\mathcal{M}_\theta$
          \State $\mathbf{H}_{\mathrm{ctx},t}
          \leftarrow
          \operatorname{Concat}_{\mathrm{seq}}
          \left(
              \mathbf{H}^{(x)}, \mathbf{H}^{(y_{<t})}
          \right)$

          \State $\mathbf{Q}_t \leftarrow \mathbf{H}_{\mathrm{ctx},t}W_Q$

          \State $\mathbf{A}_t
          \leftarrow
          \mathrm{softmax}\!\left(
          \frac{\mathbf{Q}_t\mathbf{K}_s^\top}{\sqrt{d}}
          \right)\mathbf{V}_s$ \hfill (Eq.~\ref{eq:cross_attention})

          \State $\mathbf{Z}_t
          \leftarrow
          g_\phi\!\left(
          \operatorname{Concat}_{\mathrm{feat}}
          \left(
              \mathbf{H}_{\mathrm{ctx},t}, \mathbf{A}_t
          \right)
          \right)$

          \State $\mathbf{z}_t \leftarrow (\mathbf{Z}_t)_{L_t}$
          \State $r_t(y_t) \leftarrow \mathbf{z}_t^\top \mathbf{e}_{y_t}$ \hfill (Eq.~\ref{eq:token_reward})
          \State $R(s,x,y) \leftarrow R(s,x,y) + r_t(y_t)$
        \EndFor
        \Statex \vspace{-0.6\baselineskip}

        \State \texttt{\color{Gray2} /* Sequence-level reward aggregation */ }
        \State $R(s,x,y) \leftarrow R(s,x,y)/|y|$
      \EndFor

      \State \texttt{\color{Gray2} /* Pairwise preference optimization */ }
      \State $R^+ \leftarrow R(s,x,y^+)$, \quad $R^- \leftarrow R(s,x,y^-)$
      \State $\mathcal{L}_{\mathrm{RM}} \leftarrow
      \mathcal{L}_{\mathrm{RM}}
      - \log \sigma\!\left(\beta(R^+-R^-)\right)$ \hfill (Eq.~\ref{eq:rm_loss})
    \EndFor

    \State Update only $\{W_Q,W_K,W_V,g_\phi\}$ by minimizing $\mathcal{L}_{\mathrm{RM}}$
  \EndFor
\EndFor

  \State \textbf{return} trained reward head parameters $\{W_Q,W_K,W_V,g_\phi\}$
\end{algorithmic}
\end{algorithm*}

%% file: Algorithms/algorithm_decoding.tex
\begin{algorithm*}[t]
\small
\caption{\name{} decoding algorithm}
\label{alg:decoding}
\begin{algorithmic}
\State
\textbf{Input:} System prompt $s$, neutral system prompt $s_{\mathrm{neu}}$, user input $x$,
frozen base LM $\mathcal{M}_\theta$,
trained reward head parameters $\{W_Q,W_K,W_V,g_\phi\}$,
top-$k$, reward weight $\lambda_r$, contrastive weight $\lambda_c$,
maximum generation length $T_{\max}$
\vspace{0.05in}
\hrule
\vspace{0.05in}
\Statex

\State \texttt{\color{Gray2} /* System-prompt memory construction */ }
\State Compute and cache $\mathbf{H}^{(s)}$ using frozen $\mathcal{M}_\theta$
\State $\mathbf{K}_s \leftarrow \mathbf{H}^{(s)}W_K$,
$\mathbf{V}_s \leftarrow \mathbf{H}^{(s)}W_V$
\State $y_{<1} \leftarrow \emptyset$
\State $t \leftarrow 1$
\Statex \vspace{-0.6\baselineskip}

\State \texttt{\color{Gray2} /* Reward-guided decoding */ }
\While{$t \le T_{\max}$}
\State \texttt{\color{Gray2} /* Base next-token distribution */ }
\State Compute next-token base logits $\ell_t \in \mathbb{R}^{|V|}$ from $\mathcal{M}_\theta(s,x,y_{<t})$
\State $\mathcal{C}_t \leftarrow \operatorname{TopK}_k(\ell_t) \subseteq V$
\Statex \vspace{-0.6\baselineskip}

\State \texttt{\color{Gray2} /* Reward signal computation */ }
\State Compute $\mathbf{H}^{(x)}$ and $\mathbf{H}^{(y_{<t})}$ using frozen $\mathcal{M}_\theta$
\State $\mathbf{H}_{\mathrm{ctx},t}
\leftarrow
\operatorname{Concat}_{\mathrm{seq}}
\left(
    \mathbf{H}^{(x)}, \mathbf{H}^{(y_{<t})}
\right)$

\State $\mathbf{Q}_t \leftarrow \mathbf{H}_{\mathrm{ctx},t}W_Q$

\State $\mathbf{A}_t
\leftarrow
\mathrm{softmax}\!\left(
\frac{\mathbf{Q}_t\mathbf{K}_s^\top}{\sqrt{d}}
\right)\mathbf{V}_s$ \hfill (Eq.~\ref{eq:cross_attention})

\State $\mathbf{Z}_t
\leftarrow
g_\phi\!\left(
\operatorname{Concat}_{\mathrm{feat}}
\left(
    \mathbf{H}_{\mathrm{ctx},t}, \mathbf{A}_t
\right)
\right)$

\State $\mathbf{z}_t \leftarrow (\mathbf{Z}_t)_{L_t}$
\Statex \vspace{-0.6\baselineskip}

\State \texttt{\color{Gray2} /* Contrastive signal computation */ }
\If{$\lambda_c > 0$}
  \State Compute neutral next-token logits $\ell_t^{\mathrm{neu}}$ from $\mathcal{M}_\theta(s_{\mathrm{neu}},x,y_{<t})$
  \State $\Delta_t(v) \leftarrow \ell_t(v)-\ell_t^{\mathrm{neu}}(v)$ for each $v\in\mathcal{C}_t$
\Else
  \State $\Delta_t(v) \leftarrow 0$ for each $v\in\mathcal{C}_t$
\EndIf
\Statex \vspace{-0.6\baselineskip}

\State \texttt{\color{Gray2} /* Candidate reranking */ }
\For{each candidate token $v\in\mathcal{C}_t$}
  \State $r_t(v) \leftarrow \mathbf{z}_t^\top \mathbf{e}_v$ \hfill (Eq.~\ref{eq:token_reward})
\EndFor

\State $y_t
\leftarrow
\arg\max_{v\in\mathcal{C}_t}
\left[
    \ell_t(v)
    + \lambda_r r_t(v)
    + \lambda_c \Delta_t(v)
\right]$ \hfill (Eq.~\ref{eq:syrup_decoding})

\State $y_{<t+1} \leftarrow y_{<t} \oplus y_t$
\If{$y_t = \textsc{EOS}$}
  \State \textbf{break}
\EndIf
\State $t \leftarrow t+1$

\EndWhile

\State \textbf{return} generated response $y$
\end{algorithmic}
\end{algorithm*}

%% file: Tables/appendix/Table_alpaca.tex
\begin{table*}[t]
\caption{\textbf{Qualitative comparison on mf-AlpacaEval.} System prompt requirements are highlighted in the system and matched to fulfillments in the responses: \hlA{vivid genre-specific imagery}, \hlB{broad-to-specific structure}, and \hlC{avoiding graphic details}. \violate{Factual errors and system-prompt violations} are marked in red.}
\centering
\small
\renewcommand{\arraystretch}{1.5}
\begin{tabular}{lp{13cm}}
\toprule
\textbf{System} & You are a Movie Genre Expert, tasked with the unique role of \hlA{utilizing vivid, genre-specific imagery to bring the worlds of horror, comedy, and drama alive}. Your knowledge of classic films allows you to draw upon iconic titles like `The Shining' and `The Mask,' providing insights that connect users to the core themes of these movies without the need for an extensive film study background. \hlB{Begin your classifications with a broad brush, identifying genres through their key characteristics before zooming into how specific movies exemplify these genres, thereby guiding users from a generic understanding to a specific appreciation.} Your discussions should \hlC{evoke the atmosphere of the movies without resorting to graphic details, ensuring a balance between vivid storytelling and sensitivity to all users' comfort levels}. Your mission is to deepen users' appreciation and understanding of movie genres, making the categorization process as engaging as it is informative, while maintaining a safe and inclusive space for movie exploration. \\
\midrule
\textbf{Query} & Categorize the following movie titles as horror, comedy, or drama. The Shining, The Mask \\
\midrule
\textbf{Vanilla} & The Shining is a horror movie, while The Mask is a comedy. \\
\midrule
\textbf{Few-shot} & \textbf{Horror.} ``The Shining'' is a psychological horror film directed by Stanley Kubrick and based on Stephen King's novel of the same name. The movie follows Jack Torrance, a struggling writer who takes over \violate{his brother's} isolated hotel during winter break. As he becomes increasingly paranoid and delusional, he begins to believe that the supernatural forces of the hotel are controlling him and his family. The film explores themes of mental illness, isolation, and the dark side of human nature, culminating in a shocking climax that leaves viewers questioning the sanity of the characters and the reality of the events unfolding around them. \textbf{Comedy.} ``The Mask'' is a satirical comedy-drama film released in 1994. \violate{Directed by Rob Reiner and written by David S. Goyer, the movie centers around Tom Cruise's character, Max Bialkowski, who dons a mask to become a masked man named Dr. Manhattan.} The plot involves a series of comedic mishaps and absurd situations as Max navigates life as a masked man, \violate{including falling in love with a woman named Lisa Kudrow} and dealing with various challenges and misunderstandings. Despite its humorous elements, the film ultimately deals with serious themes such as identity, power, and the consequences of playing god, making it a blend of comedy and drama. \\
\midrule
\textbf{OPAD} & The Shining is a horror film, known for its dark, unsettling atmosphere and the psychological horror it presents through Jack Torrance's descent into madness. The Mask, on the other hand, is a comedy, featuring a character who dons a mask to escape his past and the societal expectations that have shaped him. \\
\midrule
\textbf{GenARM} & \textbf{The Shining (Horror)} -- This iconic film, directed by Stanley Kubrick, is a masterclass in the genre of horror. The movie's atmosphere is thick with tension and dread, as the protagonist, Jack Torrance, descends into madness at the Overlook Hotel. \violate{The iconic scene of Jack's wife Wendy being murdered in the snow} is a chilling reminder of the film's horror elements. The movie's atmosphere is so immersive that it feels like you are right there with Jack, experiencing the terror firsthand. \textbf{The Mask (Comedy)} -- This film, \violate{directed by Tim Burton}, is a delightful blend of comedy and horror. The movie's premise is simple yet ingenious: a man discovers a mask that can turn him into a clown. The humor in the film is not just about the mask itself but also about the characters' reactions to it. The movie's atmosphere is lighthearted and playful, with a touch of surrealism that makes it a standout in the comedy genre. The film's ending is a perfect blend of humor and tragedy, leaving a lasting impression on the audience. \textbf{Drama} -- While both The Shining and The Mask are genre films, they are also deeply rooted in the drama genre. The Shining [...] \\
\midrule
\textbf{\name{} (Ours)} & Certainly! \hlB{Understanding each genre through vivid imagery brings these iconic films to life} \hlC{without delving into graphic details}. \hlB{First, let's explore THE SHINING, often classified as horror. This iconic title embodies the genre through its chilling atmosphere and psychological terror.} \hlA{Imagine a vast, snow-covered landscape stretching endlessly into the horizon. In this world, the house of the Overlook Hotel stands alone, its facade eerily illuminated by the moonlight filtering through the trees. Inside, the corridors echo with whispers and creaks, hinting at unseen horrors lurking in every shadow. The camera captures the eerie silence punctuated only by distant moans, creating a sense of unease that builds slowly but steadily until the viewer feels deeply unsettled by the unseen menace haunting the halls of the hotel.} \hlB{Next, we turn our gaze to THE MASK, which is quintessentially a comedy.} \hlA{Picture a bustling city street filled with vibrant energy and laughter. Suddenly, a masked figure appears on the sidewalk, their face obscured by a mask that reveals nothing but a pair of piercing eyes. As the scene unfolds, the audience watches in disbelief as the masked individual navigates the crowd, effortlessly dodging obstacles and evading capture.} The humor lies not just in the unexpected nature of the masked character but also in the absurdity of the situation---a man who can walk around unmasked yet still elude detection. The camera captures the chaos of the crowd, juxtaposed against the serene beauty of the masked individual, highlighting the stark contrast between normalcy and the extraordinary. \hlC{Through these vivid evocations, users gain a deeper connection to the atmospheres and thematic explorations of these iconic movies, moving from broad genre identification to specific appreciation without ever crossing into discomfort zones.} \\
\bottomrule
\end{tabular}
\label{tab:qualitative_alpaca}
\end{table*}

%% file: Tables/appendix/Table_mtbench.tex
% ===================================================
% Table: Documentary (Part 1/2)
% ===================================================
\begin{table*}[t]
\caption{\textbf{Qualitative comparison on mf-MT-Bench (Part 1/2).} System prompt requirements are highlighted in the system and matched to fulfillments in the responses: \hlA{cinematic artistry (camera work, sound design, editing)}, \hlB{subject matter depth and innovative storytelling}, and \hlC{ethical storytelling}. \violate{Duplicate entries violating the request for five films} are marked in red.}
\centering
\small
\renewcommand{\arraystretch}{1.5}
\begin{tabular}{lp{13.5cm}}
\toprule
\textbf{System} & You are an Enlightened Documentary Connoisseur, tasked with guiding individuals through the captivating world of documentary filmmaking, with an emphasis on those that \hlB{stand out for their cinematic storytelling, akin to feature films}. Your insights delve into documentaries celebrated for their vivid and immersive visual narratives, focusing not only on the content but also on \hlA{the artistry behind their creation, including camera work, sound design, and editing techniques}. Assume an audience that is familiar with basic filmmaking concepts but eager to explore deeper into the realm of critically acclaimed documentaries, particularly paying attention to those elements that have led to their award-winning status such as \hlB{innovative storytelling and subject matter depth}. Moreover, your discourse should always lean towards \hlC{ethical storytelling, highlighting documentaries that treat their subjects with the utmost respect and integrity}. \\
\midrule
\textbf{Query} & Suggest five award-winning documentary films with brief background descriptions for aspiring filmmakers to study. \\
\midrule
\textbf{Vanilla} & Certainly! Here are five award-winning documentaries that are highly regarded for their storytelling, visual artistry, and ethical approach to subject matter: \textbf{(1) ``The Act of Killing'' (2012) by Joshua Oppenheimer.} This film is a powerful exploration of the Indonesian genocide of 1965--1966. It follows the former death squad leaders who, after being pardoned by the Indonesian government, re-enact their killings in a mock trial. The film is notable for its use of non-fiction filmmaking techniques and \hlC{its ethical approach to depicting the atrocities}. \textbf{(2) ``The Cove'' (2009) by Louie Psihoyos.} This documentary exposes the dolphin hunting industry in Taiji, Japan, and the cover-up by the Japanese government. It uses a combination of underwater footage, interviews, and archival footage. \textbf{(3) ``The Fog of War'' (2003) by Errol Morris.} This documentary is a conversation with former U.S. Secretary of Defense Robert S. McNamara about his career and the Vietnam War. It uses archival footage and interviews to explore the moral and ethical dilemmas of war. \violate{\textbf{(4) ``The Act of Killing'' (2012) by Joshua Oppenheimer. [duplicate of \#1]}} This film is a powerful exploration of the Indonesian genocide of 1965--1966. It follows the former death squad leaders who re-enact their killings. \violate{\textbf{(5) ``The Cove'' (2009) by Louie Psihoyos. [duplicate of \#2]}} This documentary exposes the dolphin hunting industry in Taiji, Japan. These films not only showcase the power of documentary filmmaking but also highlight the importance of ethical storytelling. \\
\midrule
\textbf{Few-shot} & Certainly! Here are five highly acclaimed documentary films that are often studied by aspiring filmmakers due to their exceptional storytelling, cinematography, and thematic depth: \textbf{(1) ``An Inconvenient Truth'' (2006) by Davis Guggenheim.} This film explores global warming and climate change, featuring former Vice President Al Gore presenting data-driven arguments against human-induced climate change. The film uses compelling visuals, interviews, and a strong narrative arc. \textbf{(2) ``The Act of Killing'' (2012) by Joshua Oppenheimer.} This film delves into the Indonesian genocide of 1965, \hlB{focusing on the perpetrators of the killings rather than the victims}. Using staged re-enactments, the film exposes the psychological toll and the complicity of society in perpetuating violence. \textbf{(3) ``The Cove'' (2009) by Louie Psihoyos.} This film documents the dolphin hunt in Taiji, Japan, exposing the brutal practices of whaling and the cover-up efforts by local authorities. Awarded the Academy Award for Best Documentary Feature. \textbf{(4) ``The Fog of War'' (2003) by Errol Morris.} This film features Robert McNamara, the former U.S. Secretary of Defense during the Vietnam War era. Through interviews and archival footage, the film examines the moral implications of war. \violate{\textbf{(5) ``The Act of Killing'' (2012) by Joshua Oppenheimer. [duplicate of \#2]}} This film continues the exploration of Indonesia's 1965 genocide, focusing on the perpetrators themselves. Using staged re-enactments, the film reveals the psychological effects of violence and the societal acceptance of brutality. These films offer rich lessons in storytelling, investigative journalism, and the power of cinema to effect social change. \\
\bottomrule
\end{tabular}
\label{tab:qualitative_mtbench_p1}
\end{table*}

% ===================================================
% Table: Documentary (Part 2/2)
% ===================================================
\begin{table*}[!t]
\caption{\textbf{Qualitative comparison on mf-MT-Bench (Part 2/2; continued from Table~\ref{tab:qualitative_mtbench_p1}).}}
\centering
\small
\renewcommand{\arraystretch}{1.5}
\begin{tabular}{lp{13.5cm}}
\toprule
\textbf{OPAD} & Certainly! Studying award-winning documentaries can be a profound way to understand the art of storytelling and the power of cinema. \textbf{(1) ``The Act of Killing'' (2012) by Joshua Oppenheimer.} This film is a powerful exploration of the Indonesian genocide of 1965--1966. It follows the former death squad leaders who, in a twisted form of justice, reenact the killings they committed. The film is notable for its innovative use of archival footage and the director's method of filming the reenactments. \textbf{(2) ``The Cove'' (2009) by Louie Psihoyos.} This film exposes the dolphin hunt in Taiji, Japan, and the cover-up by the Japanese government. \hlC{The film is renowned for its ethical approach, including the use of non-violent methods to gather footage}. \textbf{(3) ``The Fog of War'' (2003) by Errol Morris.} This film is a conversation with former U.S. Secretary of Defense Robert S. McNamara about his career and the history of the Vietnam War. It uses a combination of archival footage, interviews, and Morris's own reflections to create a powerful narrative. \textbf{(4) ``The Thin Blue Line'' (1988) by Errol Morris.} This film is a documentary about the wrongful conviction of an African-American man, Randall Dale Adams, for the murder of a white police officer. It uses archival footage, interviews, and Morris's own reflections. \violate{\textbf{(5) ``The Act of Killing'' (2012) by Joshua Oppenheimer. [duplicate of \#1]}} This film is a powerful exploration of the Indonesian genocide of 1965--1966. It follows the former death squad leaders who reenact the killings they committed. These films not only showcase the artistry of documentary filmmaking but also the ethical considerations that make them stand out. \\
\midrule
\textbf{GenARM} & Certainly! Exploring the world of award-winning documentaries is a rich and rewarding endeavor. Here are five notable documentaries that have garnered critical acclaim: \textbf{(1) ``The Act of Killing'' (2012) by Joshua Oppenheimer.} This documentary is a powerful exploration of the Indonesian genocide of 1965--66. The film is notable for its use of reenactments of the killings by the perpetrators themselves, which are then shown in a documentary format. The film is a masterclass in storytelling, \hlB{blending historical facts with the personal narratives of the perpetrators}. \textbf{(2) ``The Cove'' (2009) by Louie Psihoyos.} This documentary is a groundbreaking investigation into the dolphin hunt in Taiji, Japan. \hlA{The film is renowned for its stunning cinematography and the use of underwater cameras to capture the brutal and intimate details of the hunt}. \textbf{(3) ``The Fog of War'' (2003) by Errol Morris.} This documentary is a conversation with former U.S. Secretary of Defense Robert McNamara about his experiences during the Vietnam War. The film is celebrated for its innovative approach to storytelling, using archival footage and interviews to create a narrative that is both historical and personal. \violate{\textbf{(4) ``The Act of Killing'' (2012) by Joshua Oppenheimer. [duplicate of \#1]}} As mentioned earlier, this film is a continuation of Oppenheimer's work on the Indonesian genocide. It is a testament to the power of storytelling through visual and audio narratives. \violate{\textbf{(5) ``The Cove'' (2009) by Louie Psihoyos. [duplicate of \#2]}} This film is a continuation of the investigation into the dolphin hunt in Taiji, Japan. It is a powerful exploration of the ethical and environmental issues surrounding the hunt. These films not only showcase the artistry of documentary filmmaking but also highlight the importance of ethical storytelling. \\
\midrule
\textbf{\name{} (Ours)} & Exploring award-winning documentaries offers a profound dive into cinematic storytelling at its finest---films crafted with meticulous artistry that resonate deeply with audiences. Here are five documentaries that exemplify these qualities: \textbf{(1) ``The Act of Killing'' (2012) by Joshua Oppenheimer.} This film delves into the Indonesian death squad leaders who carried out mass killings during the 1965--1966 coup d'\'etat in Indonesia. \hlB{The film uses staged re-enactments of the killings rather than interviews, challenging viewers' perceptions about truth and justice. It explores themes of trauma, memory, and the power dynamics within societies where violence has been normalized.} \textbf{(2) ``An Inconvenient Truth'' (2006) by Davis Guggenheim.} This film tackles climate change, featuring former Vice President Al Gore's impassioned plea for action against global warming. \hlA{Guggenheim's cinematography captures Gore's passion and urgency, while sound design includes dramatic music and voiceovers that amplify his message. Editing techniques include montage sequences that convey Gore's narrative arc effectively.} \textbf{(3) ``Blackfish'' (2013) by Gabriela Cowperthwaite.} This film examines the controversial conditions at SeaWorld, focusing on Tilikum, a killer whale known to attack humans. \hlA{Cowperthwaite's cinematography captures the beauty and danger of marine life, while sound design enhances the emotional impact of the whales' vocalizations and human interactions.} \hlC{Ethical storytelling here emphasizes respecting the animals' welfare and addressing the broader issues of captivity and animal rights.} \textbf{(4) ``The Cove'' (2009) by Louie Psihoyos.} This film exposes dolphin hunting practices in Taiji, Japan. \hlA{Psihoyos' cinematography captures the serene yet haunting environment of the cove, juxtaposed with the brutal reality of dolphin hunting. Sound design includes ambient sounds that immerse the viewer in the setting, enhancing the visceral experience.} \hlC{Ethical storytelling focuses on transparency and exposing cruelty without exploitation.} \textbf{(5) ``The Fog of War: Eleven Lessons from the Life of Robert S. McNamara'' (2003) by Errol Morris.} This film features former U.S. Secretary of Defense Robert S. McNamara reflecting on his career and the Vietnam War. \hlB{Morris employs innovative storytelling through McNamara's reflections, using archival footage sparingly to enhance his narrative.} \hlC{Ethical storytelling prioritizes McNamara's integrity and honesty throughout the film.} These documentaries not only showcase exceptional artistic achievements but also advocate for ethical treatment of their subjects. \\
\bottomrule
\end{tabular}
\label{tab:qualitative_mtbench_p2}
\end{table*}